\documentclass{article}

\usepackage{microtype}
\usepackage{graphicx}
\usepackage{subfigure}
\usepackage{booktabs} % for professional tables
\usepackage{multicol}
\usepackage{multirow}
\usepackage{caption}

\usepackage{hyperref}

\usepackage[accepted]{icml2025}

\usepackage{amsmath}
\usepackage{amssymb}
\usepackage{mathtools}
\usepackage{amsthm}

\usepackage[capitalize,noabbrev]{cleveref}

\theoremstyle{plain}
\newtheorem{theorem}{Theorem}[section]

\theoremstyle{definition}

\theoremstyle{remark}
\newtheorem{remark}[theorem]{Remark}

\usepackage[textsize=tiny]{todonotes}
\usepackage{enumitem}
\usepackage{paperpkg}

\icmltitlerunning{A Recipe for Causal Graph Regression: Confounding Effects Revisited}

\begin{document}

\twocolumn[
% \icmltitle{Causal Attention for Graph Neural Networks in Regression Task}
% \icmltitle{Rethinking Confounding Effects in Causal Graph Learning with Regression}
\icmltitle{A Recipe for Causal Graph Regression: Confounding Effects Revisited}



\icmlsetsymbol{equal}{*}

\begin{icmlauthorlist}
\icmlauthor{Yujia Yin}{equal,bu}
\icmlauthor{Tianyi Qu}{equal,sf,zju}
\icmlauthor{Zihao Wang}{hkust}
\icmlauthor{Yifan Chen}{bu}
\end{icmlauthorlist}

\icmlaffiliation{bu}{Hong Kong Baptist University}
\icmlaffiliation{hkust}{Hong Kong University of Science and Technology}

\icmlaffiliation{sf}{SF Tech}
\icmlaffiliation{zju}{Zhejiang University}

\icmlcorrespondingauthor{Tianyi Qu}{qutianyi@sf-express.com}
\icmlcorrespondingauthor{Yifan Chen}{yifanc@hkbu.edu.hk}

\icmlkeywords{Machine Learning, ICML}

\vskip 0.3in
]



\printAffiliationsAndNotice{\icmlEqualContribution} % otherwise use the standard text.

\begin{abstract}
Through recognizing causal subgraphs, causal graph learning (CGL) has risen to be a promising approach for improving the generalizability of graph neural networks under out-of-distribution (OOD) scenarios. However, the empirical successes of CGL techniques are mostly exemplified in classification settings, while regression tasks, a more challenging setting in graph learning, are overlooked. We thus devote this work to tackling causal graph regression (CGR); to this end we reshape the processing of confounding effects in existing CGL studies, which mainly deal with classification. Specifically, we reflect on the predictive power of confounders in graph-level regression, and generalize classification-specific causal intervention techniques to regression through a lens of contrastive learning. Extensive experiments on graph OOD benchmarks validate the efficacy of our proposals for CGR. 
The model implementation and the code are provided on \url{https://github.com/causal-graph/CGR}.
% in the supplementary materials and scheduled to be open-sourced upon publication. 
\end{abstract}

\section{Introduction}
\label{sec:intro}

% Graph Neural Networks (GNNs) have revolutionized the way we analyze graph-structured data by offering a powerful framework for learning representations from graphs. These models have demonstrated remarkable success across diverse applications, including social networks \cite{fan2019graph, min2021stgsn}, molecular analysis \cite{chereda2019utilizing}, traffic prediction \cite{li2021spatial, wang2020traffic}, recommendation systems \cite{shaikh2017recommendation}, and knowledge graphs \cite{sorokin2018modeling}. By leveraging a message-passing mechanism \cite{scarselli2008graph}, GNNs effectively aggregate and propagate information from neighboring nodes to capture the structural and feature-based dependencies inherent in graphs. This has led to state-of-the-art performance in key tasks such as node classification \cite{xiao2022graph}, graph classification \cite{lee2018graph, sui2022causal}, graph regression \cite{jia2020residual, pequignot2020out}, and link prediction \cite{cai2021line}.

% Graph Neural Networks (GNNs) have revolutionized the analysis of graph-structured data, offering powerful solutions across applications such as molecular analysis \cite{chereda2019utilizing}, traffic prediction \cite{li2021spatial, wang2020traffic}, and recommendation systems \cite{shaikh2017recommendation}. 
Causal graph learning (CGL) ~\citep{lin2021generative} holds particular importance due to its relevance in fields such as drug discovery ~\citep{qiao2024causal} and climate modeling ~\citep{zhao2024causal}. 
% Despite these successes, graph-level regression tasks pose unique and critical challenges in machine learning, 
% particularly in the field of causal discovery and inference\cite{5}. 
However, previous CGL studies focus on classification settings. Some of them cannot be directly extended to regression tasks, such as property prediction~\citep{rollins2024molprop}, traffic flow forecasting~\citep{li2021multistep}, and credit risk scoring~\citep{ma2024utilizing}, because the transition from finite to infinite support makes discrete labels unavailable. Graphs thus cannot be informatively grouped.
A systematical understanding of how CGL techniques should be adapted to graph-level regression is still under-explored. 

% Unlike classification tasks, where discrete labels and loss functions such as cross-entropy often help models identify causal patterns \cite{6}, 
% \yujia{regression tasks face unique challenges due to continuous targets, making it harder to directly apply some approaches.} This challenge is especially pronounced in real-world out-of-distribution (OOD) scenarios \cite{7}, where models are expected to generalize beyond the conditions encountered during training.

\begin{figure}[!t]
    \centering
    \includegraphics[trim=100 200 100 100, clip, width=1\columnwidth]{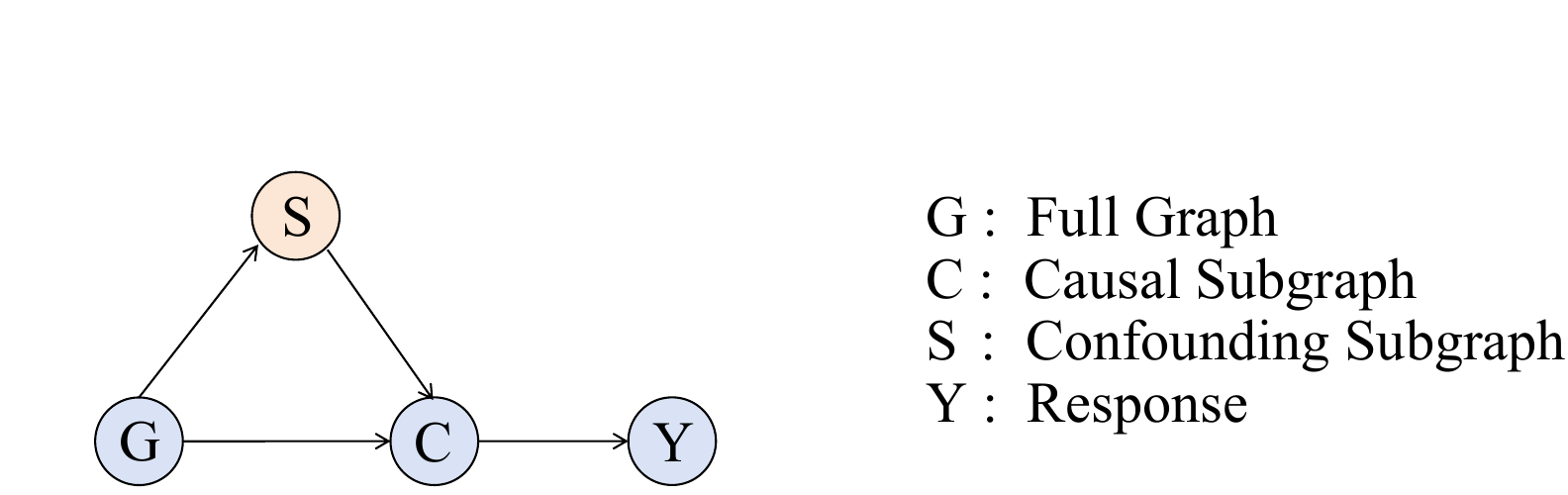}
    \caption{Structural causal model (SCM) for graph regression.
    }
    \label{fig:SCM}
\end{figure}

The core methodology of causal learning involves the identification and differentiation of causal features from confounding ones.
As shown in \Cref{fig:SCM}, causal features $C$ are those directly deciding {responses} $Y$, whereas confounding features $S$ (shorthand for ``spurious'') solely present spurious correlations. 
Therefore, understanding how causal features (as well as confounding features) and responses interact plays a central role in practical designing of causal learning methods.
From this perspective, causal graph regression (CGR) warrants specialized handling since the interaction between features and responses therein is significantly different from classification.
Furthermore, regression is in general a more challenging task than classification, and techniques working for classification, Perceptron~\citep{rosenblatt1958perceptron} for example, may not apply to regression.

% indirectly associated, often through spurious correlations.
% \tyq{check the definition of the causal effects. The description is not professional enough.}

Specifically in CGL, the identification of causal subgraphs is seemingly transferable since this step, explicitly or implicitly, relies on the calculation of mutual information and is compatible with both settings (c.f.\ \Cref{sec:prelim-gib}).
However, the empirical performance of this vanilla adaptation on regression tasks 
% (the GSAT method is indeed outstanding)
% (denoted as XXXX \yc{what is the name of the vanilla method simply using GIB on regression?}) 
is dwarfed by empirical risk minimization~\citep[ERM]{vapnik1991principles} w.r.t.\ least squares loss (see the results in \Cref{sec:exp-good,sec:exp-rood}).
% although CGL studies bloom and bring unprecedented empirical success, CGR benefits little from the development due to the inapplicability of some CGL techniques.
% Particularly, 

To crack CGR, we revisit the processing of confounding effects, which conceptually constitutes causal graph learning along with causal subgraph identification as shown in \Cref{fig:SCM}. 
% and investigate its appropriate adaptation to regression.
% In addition to the clear distinction in loss functions for learning from supervised signals (for example, cross-entropy loss in classification and least squares loss in regression), the technical processing of confounding effects substantially differs in the two settings,
% which hinders the development of CGR.
% In addition, challenges in regression tasks spotlight the 
% previous methods pay much attention to identification of causal subgraphs
%  receive less attention than causal subgraphs in previous methods.
% ; considering the two factors together constitute causal diagrams of interest 
% Previously, 
% A key yet under-explored issue in graph regression is the influence of confounding features. 
Existing CGL methods, such as CAL~\citep{sui2022causal} and DisC~\citep{fan2022debiasing}, are built on a strong assumption that confounding subgraphs contain strictly no predictive power.
We reflect on this assumption and speculate it is hardly practical due to the contradiction with real-world observations:
% focus on enhancing generalization by identifying causal subgraphs with strong predictive power 
% . However, these approaches often treat confounding subgraphs as noise, which can severely compromise generalization. 
% In classification tasks, the impact of this limitation is mitigated by the discrete and limited nature of labels. In contrast, regression tasks amplify the issue, as reliance on spurious correlations becomes more pronounced.
% \tyq{No support, any reference? Statistical papers?} 
% For example, 
in molecular property prediction, for example, molecular weight is noncausal to toxicity while does exhibit strong correlations.
% Models relying on these correlations often fail in out-of-distribution (OOD) environments where such relationships no longer hold. 
% Addressing this limitation requires a fundamental shift in causal discovery for graph regression.

In this work, we develop an enhanced graph information bottleneck (GIB) loss function, 
which no longer takes the strong assumption.
Moreover, some confounding effect processing techniques, such as backdoor adjustment~\citep{sui2022causal,sui2024enhancing} and counterfactual reasoning~\citep{guo2025counterfactual},
heavily rely on discrete label information and cannot be adapted to regression at all. 
% the label-based causal intervention is not suitable for regression tasks. 
% Existing causal intervention techniques, such as backdoor adjustment~\citep{XXX} and counterfactual reasoning~\citep{XXXX}, 
% have achieved remarkable success in graph classification tasks. 
% However, these methods heavily rely on discrete label information and classification-specific loss functions
% \tyq{should we mention practically. Theoretically, there is no such issue. And dose it rely on the classification-specific loss function?}, 
% making them unsuitable for regression tasks. 
We follow the principle of those methods and generalize it from class separation to instance discrimination;
the discrimination principle aligns with the philosophy of contrastive learning (CL) and CL techniques are therefore leveraged to tackle CGR in our proposal.

Following the intuition, we develop a new framework for causal graph regression, which spotlights the confounding effects within. 
In summary, 
% that integrates causal attention mechanisms with regression-specific learning objectives. 
our contributions are as follows:
% \begin{itemize}[topsep=0em, leftmargin=*]
\begin{itemize}[topsep=0em, leftmargin=*]
    \item To the best of our knowledge, we are the first to explicitly consider the predictive role of confounding features in graph regression tasks, a critical yet overlooked aspect in graph OOD generalization.
    \item We introduce a new causal intervention approach that generates random graph representations by leveraging a contrastive learning loss to enhance causal representation, outperforming label-dependent methods.
    \item Extensive experiments on OOD benchmarks demonstrate that our method significantly improves generalization in graph regression tasks.
\end{itemize}
% By addressing these critical gaps, our work provides a foundation for advancing causal discovery in graph regression tasks, bridging the methodological divide between classification and regression in causal inference for GNNs. This not only improves the generalization of GNNs in real-world applications but also deepens our understanding of causal relationships in complex graph-based data.

% \begin{figure*}[h!]
%     \centering
%     \includegraphics[trim=0 0 0 0,clip,width=1.0\textwidth]{fig/fig1.png}
%     \caption{Illustration of the proposed model architecture(draft).}
%     \label{fig:fig1}
% \end{figure*}

\section{Related Work}

Out-of-distribution (OOD) challenges in graph learning has drawn significant attention, particularly in methods aiming to disentangle causal and confounding factors~\citep{ma2024survey}. 
Existing approaches can be broadly categorized into invariant learning ~\citep{wu2022handling}, causal modeling ~\citep{sui2024enhancing}, and stable learning ~\citep{li2022ood}.

\textbf{Invariant learning} focuses on identifying features that remain stable across different environments, filtering out spurious correlations in the process. 
While not explicitly grounded in causal reasoning, prior studies ~\citep{wang2022unified, mitrovic2020representation} have highlighted its inherent connection to causality. 
Methods in invariant learning, such as CIGA ~\citep{chen2022learning}, GSAT ~\citep{miao2022interpretable}, and GALA ~\citep{chen2024does}, aim to learn invariant representations by isolating causal components. 

However, these approaches are typically designed for classification tasks, limiting their out-of-distribution (OOD) generalization capability in regression settings. Post-hoc methods, such as PGExplainer ~\citep{luo2020parameterized} and RegExplainer ~\citep{zhang2023regexplainer}, attempt to discover invariant subgraphs after training. However, these methods fail to equip the model with the ability to learn invariant representations during the training process.

\textbf{Causal modeling} leverages structural causal models (SCMs) to improve the performance of graph neural networks (GNNs) on out-of-distribution (OOD) data. These approaches incorporate various traditional causal inference techniques, such as backdoor adjustment (e.g., CAL ~\citep{sui2022causal}, CAL+ ~\citep{sui2024enhancing}), frontdoor adjustment (e.g., DSE ~\citep{wu2022deconfounding}), instrumental variables (e.g., RCGRL ~\citep{gao2023robust}), and counterfactual reasoning (e.g., DisC ~\citep{fan2022debiasing}). By simulating causal interventions through supervised training, these methods aim to achieve OOD generalization. However, they often disregard the predictive potential of confounding features, which hinders effective disentanglement. Moreover, the supervised loss functions tailored for classification tasks are not easily adaptable to regression problems, as the inherent complexity of regression introduces additional challenges.

\textbf{Stable learning} aims to ensure consistent performance across environments by reweighting samples or balancing covariate distributions. For example, StableGNN ~\citep{fan2023generalizing} employs a regularizer to reduce the influence of confounding variables. However, such methods often rely on heuristic reweighting strategies, which may not fully disentangle causal from confounding factors.

In addition to graph-based approaches, traditional machine learning methods have also explored causality in regression tasks. For instance, \citet{pleiss2019neural} observed that causal features tend to concentrate in a low-dimensional subspace, whereas non-causal features are more randomly distributed. Similarly, \citet{amini2020deep} proposed a framework for learning continuous targets by placing an evidence prior on a Gaussian likelihood function and training a non-Bayesian neural network to infer the hyperparameters of the evidence distribution. These methods highlight the potential of leveraging causal insights for improved regression performance.

\section{Preliminaries and Notations}

% \subsection{}
Along this paper, we denote a graph $G$ as $\left(\mtx A, \mtx X \right)$. 
Here, $\mtx A \in \{0, 1\}^{n \times n}$ is the adjacency matrix indicating connectivity among $n$ nodes ($\mtx A_{ij} = 1$ if nodes $i$ and $j$ are connected, otherwise 0); 
$\mtx X \in \mathbb{R}^{n \times d}$ is the node feature matrix, where each row $\mtx X_i$ represents the $d$-dimensional feature vector of node $i$. 
The regression task in graph learning is to learn a function $f: G \mapsto y$, where $y \in \mathbb{R}$ denotes the response for the graph $G$.

\subsection{Causal Graph Learning}
\label{sec:prelim-cgl}

In causal graph learning, a graph $G$ can be split into a \textbf{causal subgraph} $C$ and a \textbf{confounding subgraph} $S$.
This process is non-trivial and our proposed paradigm will hinge on the output of this process.
We follow the definition in \citet{sui2022causal} and first introduce the construction of the causal subgraph $C$: 
\begin{align}
C \defeq (\mtx M_\text{edge} \odot \mtx A, \mtx M_\text{node} \cdot \mathbf X),
\label{eqn:causal-sub}
\end{align}
where the mask matrix $\mtx M_\text{edge} \in [0, 1]^{n \times n}$ and the diagonal matrix $\mtx M_\text{node}$ (whose diagonal elements are in $[0, 1]$) will filter out the non-causal nodes and edges. The confounding subgraph is then the ``complement'': $S \defeq G - C$.

In our framework, these masks $\mtx M_\text{edge}$ and $\mtx M_\text{node}$ are not pre-defined. Instead, they are learnable soft masks, generated by MLPs conditioned on the representations of $G$. The parameters of these MLPs are optimized end-to-end as part of the overall model training, enabling the model to autonomously learn how to construct $C$ and $S$. Further architectural details are provided in Appendix~\ref{appendix:framework} and illustrated in~\Cref{fig:fig1}.

Notably, mutual information plays an essential role in CGL, and we introduce its calculation, exemplified by the mutual information between the hidden embedding vectors (learned by a graph neural network) of the causal subgraphs and the original graphs, as follows:
\begin{equation}
I(C; G) \defeq \mathbb{E}_{C, G} \left[ \log {p(C \mid G)}/{p(C)} \right],
\end{equation}
where we follow the convention in CGL literature and abuse the notation $C, G$ to represent a random variable following the \textbf{underlying distribution of embedding pairs} $H_{g, i}$'s and $H_{c, i}$'s.
In particular, those hidden embeddings are assumed Gaussian and the joint distribution can thus be well-estimated by sample embedding pairs.
We refer readers interested to \citet[Appendix~A]{miao2022interpretable} for more details.
Moreover, the computation/approximation of the mutual information terms is a crucial component in causal graph learning, while still under-explored for CGR;
we will dissect the computation of our proposed terms in \Cref{sec:Disentanglement Mechanism} through deriving the variational bounds.

\subsection{Graph Information Bottleneck}
\label{sec:prelim-gib}

The information bottleneck~\citep[IB]{tishby2000information, tishby2015deep} principle aims to 
% learning representations that are both compact and predictive. It achieves this by 
balance the trade-off between preserving the information necessary for prediction and discarding irrelevant redundancy. 
Specifically, IB suggests to maximize $I(Z; Y)$ while minimizing $I(Z; X)$ for regular data compression, where $Z$ is the compressed representation, $X$ is the input, and $Y$ is the response.

Graph information bottleneck (GIB)~\citep{wu2020graph} extends the IB principle to graph-structured data, facilitating the identification of subgraphs that are most relevant for predicting graph-level responses. 
% GIB focuses on extracting subgraphs that retain essential predictive information while excluding redundant components. 
By minimizing the mutual information $I(C; G)$ between the extracted causal subgraph $C$ and the original graph $G$, 
GIB reduces redundant information.
% , which may include confounding factors. 
However, GIB alone does not guarantee the extraction of a purely causal subgraph, as isolating causal effects requires additional interventions ~\citep{miao2022interpretable,chen2022learning}. 
% Nonetheless, this process can benefit the identification of structural patterns that are more predictive and contribute to disentangling causal relationships from confounding influences.

Formally, the GIB objective is expressed as:
\begin{equation}
-I(C; Y) + \alpha I(C; G), 
% \quad C \in \mathcal{G}_{\text{sub}}(G),
\end{equation}
where $I(C; Y)$ quantifies the predictive information retained by $C$ (and thus needs to maximize). 
$I(C; G)$ serves as a regularizer to exclude irrelevant details from the original graph;
the parameter $\alpha$  controls the trade-off between information preservation and compression. 
% To operationalize these mutual information terms, the graph and subgraph are typically represented as embeddings learned by a graph neural network. 

% By embedding the graphs into a latent space, these mutual information terms are made computationally tractable, allowing the GIB framework to be applied in practical settings. A notable strength of GIB is its flexibility, as it does not impose rigid constraints such as fixed subgraph size or connectivity, making it applicable to diverse graph learning tasks. By focusing on task-relevant substructures, GIB enhances interpretability and robustness in graph neural networks while naturally aligning with causal reasoning. This makes it a powerful tool for learning representations that are both effective and meaningful.

\subsection{Causal Intervention in GNNs}

% The Structural Causal Model (SCM) provides a framework for analyzing causal relationships in Graph Neural Networks (GNNs). 
We borrow the structural causal model (SCM) diagram in \Cref{fig:SCM} to illustrate the causal intervention techniques. 
As shown in \Cref{fig:SCM}, the graph $G$ decides both the causal subgraph $C$ and the confounding subgraph $S$, and the former $C$ affects the prediction of response $Y$. In more detail,
\begin{itemize}[topsep=-0.2em, itemsep=-0.4em, leftmargin=*]
  \item $C \leftarrow G \rightarrow S$: Graph data $G$ encodes both $C$, which directly impacts $Y$, and $S$, which introduces spurious correlations.
  \item $S \rightarrow C \rightarrow Y$: The causal feature $C$ has the potential to predict $Y$ not only directly but also indirectly through its influence along this backdoor path $S \rightarrow C \rightarrow Y$.
\end{itemize}
% In causal inference, confounder $S$ causes spurious correlations, preventing direct associations from reflecting true causal relationships. The backdoor adjustment method addresses this issue by estimating $P(Y|C,S)$, where $P(Y|C,S)$ represents the conditional distribution of response $Y$ given $C$ and $S$. By stratifying over $S$, backdoor adjustment computes the interventional effect $P(Y|\text{do}(C))$, effectively eliminating confounder. 
In causal inference, confounder $S$ incurs spurious correlations, preventing the discovery of underlying causality. 
To address this issue, backdoor adjustment methods focus on the interventional effect $P(Y|\text{do}(C))$, and suggest to estimate it by stratifying over $S$ and calculating the conditional distribution $P(Y|C,S)$~\citep{pearl2014interpretation,sui2024enhancing}.

% $P(Y|C,S)$, where $P(Y|C,S)$ represents the conditional distribution of response $Y$ given $C$ and $S$. By stratifying over $S$, backdoor adjustment computes the interventional effect $P(Y|\text{do}(C))$, effectively eliminating confounder. 

% In regression tasks, $S$ can provide some predictive power but undermines generalization, especially under distribution shifts. Causal intervention addresses this by cutting off the backdoor path, producing robust and generalizable predictions.

\begin{figure*}[t]
    \centering
    \includegraphics[trim=710 0 750 50,clip,width=0.99\textwidth]{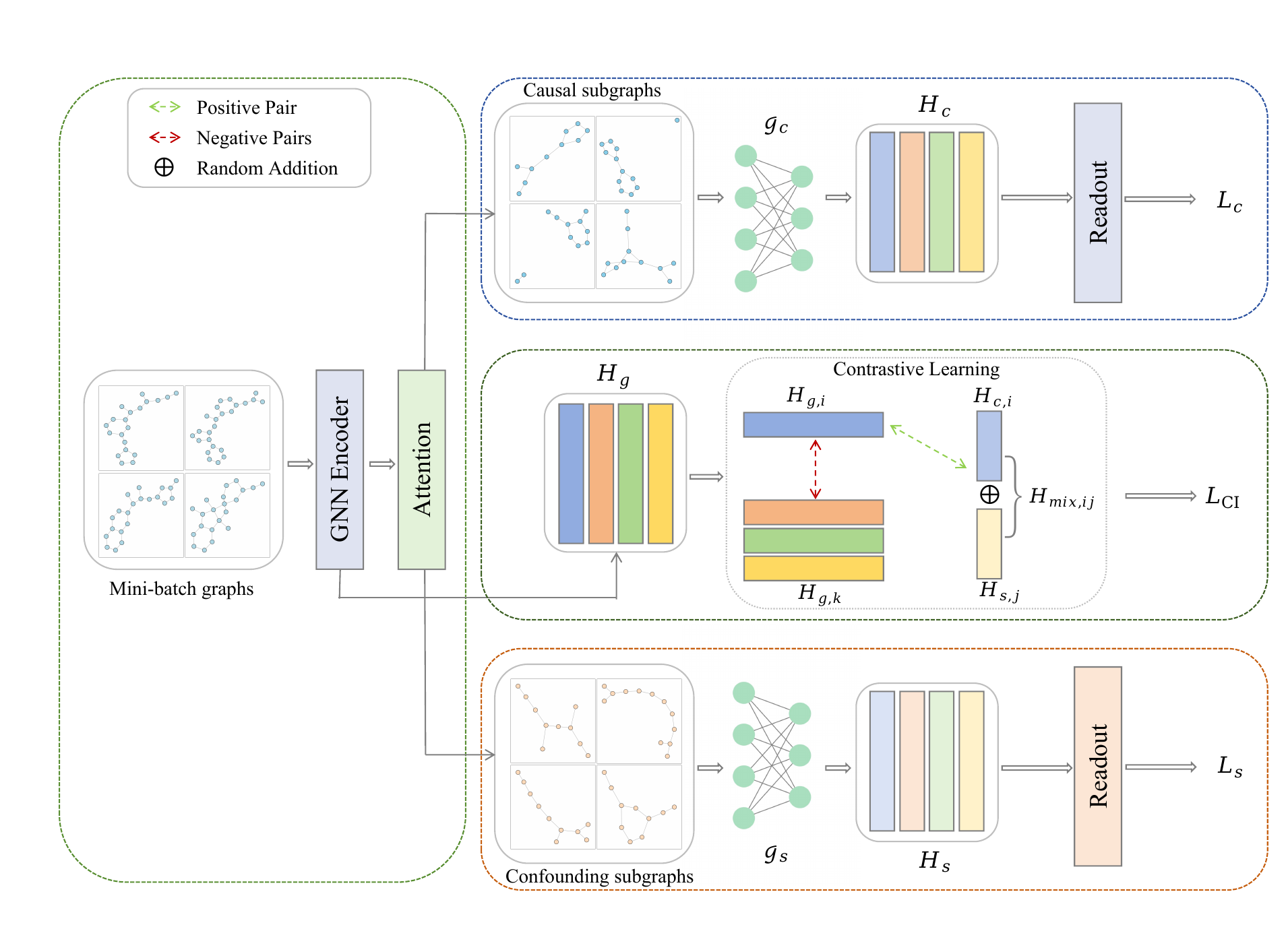}
    \caption{Given a mini-batch of graphs, 
    (1) the GNN encoder computes the graph embeddings $H_g$, 
    and an attention layer generates soft masks to extract causal and confounding subgraphs. 
    (2) GNN $\mathcal{G}_{c}$ processes the causal subgraph $C$, generates its representation $H_c$, and employs readout to predict responses; it is optimized with causal subgraph loss $L_c$. 
    (3) GNN $\mathcal{G}_{s}$, sharing parameters with $\mathcal{G}_{c}$, processes the confounding subgraph $S$, generates $H_s$, and applies readout for prediction; it is optimized with confounding subgraph loss $L_s$. 
    (4) For causal intervention, contrastive learning guides the process. 
    Given a graph $H_{g,i}$, the positive sample is a mixed graph $H_{\mathrm{mix},ij}$ from random addition, while any other graph $H_{g,k}$ serves as the negative sample. The causal intervention loss $L_{\mathrm{CI}}$ is used accordingly.
    }
    \label{fig:fig1}
% \vspace{-2em}
\end{figure*}

\section{Revisiting Confounding Effects for CGR}

In this section, we present 
% Reg-OOD\tyq{change name}, 
a causal graph regression paradigm that integrates an enhanced graph information bottleneck (GIB) objective with causal discovery,
reshaping the processing of confounding effects in CGL. 

\subsection{Overview}
\label{sec:overview}
% Given a mini-batch of graphs $\mathbb{G}=\left\{G_1,\cdots,G_n\right\}$\tyq{the following does not need a batch. The operation is for a single graph}, we employ an attention mechanism to distinguish between causal and confounding subgraphs with each graph. Particularly, 

We first provide an overview of how graph inputs are turned into regression outputs.
As shown in \Cref{fig:fig1}, we follow the framework of \citet{sui2024enhancing} and first encode graph embeddings $H_{g, i}$'s using a GNN-based encoder.
Attention modules are then adopted to generate soft masks for extracting causal and confounding subgraphs (c.f.\ \Cref{eqn:causal-sub}). 
% and employs an attention module to generate soft masks, extracting causal and confounding subgraphs. 
% Specifically, node and edge attention scores serve as soft masks are estimated using separate MLPs, enabling the model to focus on key structural components while maintaining differentiability. 
% The masked graphs are then decomposed into causal ($C$) and confounding ($S$) subgraphs. 
These subgraphs are processed through two GNN modules ($\mathcal{G}_c$ and $\mathcal{G}_s$) with shared parameters to extract causal ($H_{c, i}$'s) and confounding ($H_{s, i}$'s) representations, which are passed through distinct readout layers for regression.

The optimization features an enhanced graph information bottleneck (GIB) loss $L_\mathrm{GIB}$, comprising the causal part $L_c$ and the confounding part $L_s$, to disentangle causal signals (c.f.\ \Cref{sec:Disentanglement Mechanism}). 
Also, counterfactual samples ($H_{\mathrm{mix},ij}$) are generated by randomly injecting confounding representations into causal ones; 
unsupervised learning is then performed, guided by contrastive-learning-based causal intervention loss $L_\mathrm{CI}$ (c.f.\ \Cref{sec:intervention}).
% , improving out-of-distribution generalization in graph regression tasks. 
More implementation details of the overall framework are deferred to \Cref{appendix:framework}.
% due to space limit.

% We next introduce a series of loss functions to disentangle their representations and guide the causal learning process, enabling effective out-of-distribution generalization on graph regression tasks.

\subsection{Enhanced GIB Objective}
\label{sec:Disentanglement Mechanism}

CGL adopts the GIB objective to extract subgraphs that retain essential predictive information while excluding redundant components~\citep{zhang2023regexplainer}, which aligns with the disentanglement of causal subgraph $C$ and confounding subgraph $S$ in CGL. 
Original GIB assumes the confounding subgraph $S$ is pure noise and cannot predict the response $Y$~\citep{chen2022learning}, 
while as we discussed in \Cref{sec:intro} $S$ may still contain information that is predictive of the response $Y$. 
In its current form, the GIB framework overlooks this aspect, causing the model to allocate all $Y$-relevant information to $C$ 
% while $S$ is treated as noise, 
and to potentially lose meaningful content.

This limitation leads to incomplete causal disentanglement, which impacts the generalization of models to out-of-distribution (OOD) settings. 
To overcome this issue, we propose an enhanced GIB loss function that takes the predictive roles of both $C$ and $S$ into consideration. 
By introducing mutual information terms on $S$ during optimization, we avoid overburdening $C$ with all relevant information, 
and consequently enable a more precise disentanglement. 

Overall, our enhanced GIB objective is defined as follows:
\begin{align}
    -I(C; Y) + \alpha I(C; G) -\beta I(S;Y),
    \label{eq: enhanced GIB}
\end{align}
which formally extends the original GIB objective by introducing a confounder-related term $I(S;Y)$ to capture the predictive capacity of $S$, along with a parameter $\beta$. 
% Unlike the causal subgraph $C$, we do not explicitly constrain the mutual information $I(S;G)$ 
In particular, we intentionally exclude the $I(S;G)$ term because, in the SCM diagram of \Cref{fig:SCM}, $S$ primarily introduces shortcut rather than directly encoding causality;
overly imposing structural regularization on $S$ can disrupt disentanglement and  lead to suboptimal separation between $C$ and $S$.
% Simply minimizing $I(S;G)$ or imposing other structural regularization may disrupt disentanglement, leading to suboptimal separation between $C$ and $S$. Therefore, our enhanced GIB objective preserves the predictive signal within $S$ while preventing it from dominating the causal discovery.
Notably, the conceptual objective \eqref{eq: enhanced GIB} is incomputable in practice. 
We devote the remainder of this subsection to the practical computation of \Cref{eq: enhanced GIB} for CGR.

\textbf{Variational bounds for approximating $I(C; G)$.}
The mutual information $I(C; G)$ 
% quantifies the dependency between $C$ and $G$ and 
is mathematically defined based on the marginal distributin $p(C)=\sum_G p(C|G)p(G)$. 
Since $p(C)$ is intractable, a variational distribution $q(C)$ is introduced and induces an upper bound:
\begin{equation}
I(C; G) \leq \mathbb{E}_{p(G)} \big[ \mathrm{KL}\big(p(C \mid G) \| q(C)\big) \big]. \label{eq:mi_upper_bound}
\end{equation}
\begin{comment}
% However, computing the marginal distribution $p(C) = \sum_G p(C \mid G) p(G)$ is intractable, to overcome this challenge, we approximate $p(C)$ with a variational distribution $q(C)$. Substituting $q(C)$ into Eq.~\eqref{eq:mutual_info}, we reformulate $I(C; G)$ as:
% \begin{equation}
% I(C; G) = \mathbb{E}_{p(C, G)} \left[ \log \frac{p(C \mid G)}{q(C)} \right] - \mathrm{KL}\big(p(C) \| q(C)\big). \label{eq:variational_mi}
% \end{equation}

% The KL divergence term $\mathrm{KL}\big(p(C) \| q(C)\big)$ is non-negative, providing an upper bound for $I(C; G)$:
% \begin{equation}
% I(C; G) \leq \mathbb{E}_{p(G)} \big[ \mathrm{KL}\big(p(C \mid G) \| q(C)\big) \big]. \label{eq:mi_upper_bound}
% \end{equation}
\end{comment}
To efficiently compute the KL divergence in \Cref{eq:mi_upper_bound}, we follow the literature~\citep{chechik2003information, kingma2013auto} and assume that $p(C \mid G)$ and $q(C)$ are multivariate Gaussian distributions:
\begin{equation}
p(C \mid G) = \mathcal{N}(\mu_{\phi}(G), \Sigma_{\phi}(G)), \quad q(C) = \mathcal{N}(0, I),
\end{equation}
where $\mu_{\phi}(G)$ and $\Sigma_{\phi}(G)$ are the mean vector and covariance matrix estimated by GNNs. 
To simplify computation and stabilize training, 
we further assume $\Sigma_{\phi}(G)$ is an identity matrix, 
% i.e., $\Sigma_{\phi}(G) = \mathrm{diag}(\sigma^2_{\phi}(G))$, and further fix it as $I$, 
removing the need to learn covariance parameters. This simplification is not only practical but also theoretically justified, as any full-rank covariance can be whitened without loss of generality~\citep[Appendix A]{chechik2003information}.
$\mathrm{KL}\big(p(C \mid G) \| q(C)\big)$ then reduces to:
\begin{equation}
\begin{aligned}
& \frac{1}{2} \Big[ \mathrm{tr}(\Sigma_{\phi}(G)) + \|\mu_{\phi}(G)\|^2 - d - \log \det \Sigma_{\phi}(G) \Big] \\
% =& \frac{1}{2} \sum_{i=1}^d \big( \sigma_{\phi,i}^2(G) + \mu_{\phi,i}^2(G) - 1 - \log \sigma_{\phi,i}^2(G) \big)\\
=& \frac{1}{2}\|\mu_{\phi}(G)\|^2.
\end{aligned}
\label{eq:kl_simplified}
\end{equation}
where $d$ is the dimensionality of $C$.
Further substituting \Cref{eq:kl_simplified} into \Cref{eq:mi_upper_bound}, we obtain an upper bound for $I(C; G)$:
\begin{equation}
I(C; G) \leq \frac{1}{2} \mathbb{E}_{p(G)} \big[\|\mu_{\phi}(G)\|^2\big], 
\label{eq:mi_final}
\end{equation}
which serves as an easy-to-compute proxy for $I(C; G)$.

\textbf{Variational bounds for approximating $I(C;Y), I(S;Y)$.}
% Building on the variational approximation of $I(C; G)$, we extend the discussion to estimating $I(C; Y)$, which measures the dependency between the causal subgraph $C$ and the target variable $Y$. This metric reflects the predictive power of $C$ for predicting $Y$. 
We first recall $I(C; Y)$ mathematically reads:
\begin{equation}
  I(C; Y) = H(Y) - H(Y \mid C),  
\end{equation}
where $H(Y)$ denotes the entropy of $Y$, representing the overall uncertainty in the target variable. 
Since $H(Y)$ remains constant, maximizing $I(C; Y)$ reduces to minimizing the conditional entropy $H(Y \mid C)$, given by:
\begin{equation}
  H(Y \mid C) = -\mathbb{E}_{C, Y}[\log p(Y \mid C)].  
\label{eqn:conditional entropy}
\end{equation}
% \yc{@Tianyi. check the expectation}
% \tyq{checked}%https://en.wikipedia.org/wiki/Conditional_entropy
% To make $H(Y \mid C)$ computationally tractable, the causal subgraph $C$ is processed
The computation of $H(Y \mid C)$ is supposed to hinge on the hidden embeddings $H_{c, i}$'s produced by a GNN $\mathcal{G}_c$ (see \Cref{sec:overview});
% Notably in practice, we using a graph neural network $\mathcal{G}_c$ to produce representation $H_c$. This representation encodes the relevant features of $C$ for predicting $Y$. 
we model the conditional distribution $p(Y \mid H_c)$ as a Gaussian distribution:
\begin{equation}
   p(Y \mid H_c) = \mathcal{N}(Y; \mu_{(c)}, \sigma^2_{(c)}), 
\end{equation}
where $\mu_{(c)}$ and $\sigma^2_{(c)}$ represent the scalar conditional mean and variance of $Y$ (estimated by networks) given a causal subgraph representation $H_c$. 
The probability density function for this Gaussian is:
\begin{equation}
    p(Y \mid H_c) = \frac{1}{\sqrt{2\pi\sigma^2_{(c)}}} \exp\left(-\frac{(Y - \mu_{(c)})^2}{2\sigma^2_{(c)}}\right).
    \label{eq:p(Y|hc)}
\end{equation}
Substituting \Cref{eq:p(Y|hc)} into \Cref{eqn:conditional entropy}, we can further approximate $H(Y \mid C)$ through empirical data:
% \begin{equation}
%     H(Y \mid C) = \mathbb{E}_{p(c, Y)}\left[\frac{(Y - \mu(c))^2}{2\sigma^2(c)} + \frac{1}{2} \log(2\pi\sigma^2(c))\right].
% \end{equation}
\begin{equation}
\frac{1}{N} \sum_{i=1}^N \left[\frac{(Y_i - \mu_{(c), i})^2}{2\sigma^2_{(c), i}} + \frac{1}{2} \log(2\pi\sigma^2_{(c), i})\right], 
\end{equation}
where $N$ represents sample size, $Y_i$ is the target response for the $i$-th sample.
and $\mu_{(c), i}$ and $\sigma^2_{(c), i}$ are the corresponding mean and variance of $Y$ given $H_{c, i}$. 

If a constant conditional variance (i.e., $\sigma^2_{(c)} = 1$) is assumed, a choice adopted for stability and aligning with approaches in~\citep{nix1994learning, yu2024cauchy}, then $I(C;Y)$ (or, equivalently, $-H(Y\mid C)$) reduces to the least squares loss:
\begin{align}
& -\frac{1}{N} \sum_{i=1}^N \left[\frac{(Y_i - \mu_{(c), i})^2}{2\sigma^2_{(c), i}} + \frac{1}{2} \log(2\pi\sigma^2_{(c), i})\right] \nonumber \\
\propto& -\frac{1}{N} \sum_{i=1}^N (Y_i - \mu_{(c), i})^2,
\end{align}
which turns to the \textbf{causal subgraph objective} $L_\text{CP}$.
% This framework integrates the Gaussian assumption into mutual information estimation, allowing for optimization of the dependency between $C$ and $Y$. 

Similarly, the mutual information $I(S; Y)$ 
% between the confounding subgraph $S$ and $Y$ 
can induce the \textbf{confounding subgraph objective} 
% be estimated analogously. Assuming a constant variance, its predictive loss function is:
\begin{equation}
  L_\text{SP} \propto -\frac{1}{N} \sum_{i=1}^N (Y_i - \mu_{(s), i})^2. 
\end{equation}
Empirically, we employ two independent readout layers to compute the causal and confounding subgraph mean $\mu_{(c), i}$'s and $\mu_{(s), i}$'s. 

In summary, our enhanced GIB objective can be decomposed into two distinct loss components: the causal subgraph loss $L_{c}(G,C,Y)=-I(C;Y)+\alpha I(C;G)$ and the confounding subgraph loss $L_{s}(S,Y)=-I(S;Y)$.
% In our design, we assign $L_{c}=-I(C;Y)+\alpha I(C;G)$ and $L_{s}=-I(S;Y)$. 
% Based on this setup, 
The complete enhanced GIB objective we propose is:
\begin{align*}
    L_\text{GIB} &= L_{c} + \beta L_{s} \\
    &= -I(C;Y)+ \alpha I(C;G) - \beta I(S; Y),
\end{align*}
and in practice we use $- L_\text{CP} + \alpha \mathbb{E}_{p(G)} \big[ \|\mu_\phi(G)\|^2 \big]  - \beta L_\text{SP}$.

\subsection{Causal Intervention}
\label{sec:intervention}

To further strengthen causal learning in CGR, we introduce a causal intervention loss and reshape the processing of confounding effects therein. 
% According to \citet{miao2022interpretable, miao2022interpretable2}, introducing randomness during causal training is crucial for achieving OOD generalization. 
% Previously, GSAT~\citep{miao2022interpretable} selected task-relevant subgraphs by independently sampling nodes or edges using a Bernoulli distribution. 
% However, this independent sampling process may disrupt the original graph structure, affecting its connectivity and topological characteristics, which in turn impacts model performance. 
In general, our approach injects randomness at the graph level by randomly pairing confounding subgraphs with target causal subgraphs from the entire dataset. By generating counterfactual graph representations through the random combination of these subgraphs, we effectively implement causal intervention.

% \yujia{
This strategy can be understood as an implicit realization of backdoor adjustment~\citep{pearl2014interpretation} in the representation space. 
In existing research on graph classification tasks~\citep{fan2022debiasing,sui2024enhancing}, 
causal intervention is typically modeled by predicting $P(Y|C,S)$ through intervened graphs, 
adjusting for causal effects by comparing predictive distributions under different confounding conditions. 
However, in regression tasks, $Y$ is a continuous variable, and directly modeling $P(Y|C,S)$ becomes significantly more challenging. 
To overcome this, we follow the spirit of contrastive learning to get rid of the reliance on explicit labels.
% } 
% Instead of explicitly modeling predictive distributions, we construct counterfactual graphs and minimize the representation difference between them and the original graphs. This approach enables the model to implicitly de-confounder, leading to more robust causal learning. Ultimately, our method not only mitigates confounding effects in regression tasks but also improves OOD generalization in GNN models.

In more detail, following \citet{sui2022causal}, we use a random addition method to pair the confounding subgraph with the target causal subgraph %to simulate counterfactual.the representation of paired graph 
, which gives $H_{\mathrm{mix}}$:
\begin{equation}
    H_{\mathrm{mix},ij} = H_{c,i} + H_{s,j}.
\end{equation}
%Given the regression nature of our task, directly
Comparing the predictions of $H_{\mathrm{mix}}$ with the original graph's labels, as shown in \citet{sui2022causal}, can inadvertently force the mixed graph to discard all confounding effects, 
thereby nullifying the intended causal disentanglement. 

% However, there is no 
To mitigate this issue, we suggest learning causal representations through contrastive learning. 
Specifically, 
%our approach ensures that 
the causal subgraph, when combined with different confounding subgraphs, consistently produces mixed graph representations that are aligned with the original graph representation. This formulation enables the model to learn causal subgraphs that are invariant across varying confounders, and to avoid the causal subgraphs boiled down to non-informative ones.

To achieve this, we propose a causal intervention loss guided by contrastive learning. 
Specifically, the method aligns the representation of the original graph with that of its corresponding random mixture graph, while simultaneously ensuring that representations of unrelated graphs remain distinct. 
In implementation, draw inspiration from the InfoNCE loss ~\citep{oord2018representation}, we treat $H_{g}$ and $H_{\mathrm{mix}}$ from the same causal subgraph as positive pairs,  and $H_{g}$ with representations of other graphs within the batch as negative pairs. Formally, the mixed graph contrastive loss is defined as:
\begin{equation}
    L_{\mathrm{CI}}= -\frac{1}{B}\sum_{i=1}^B \log \frac{\exp (\mathrm{sim}( H_{g,i}, H_{\mathrm{mix},ij})}{\sum_{\substack{k=1, k \neq i}}^B \exp (\mathrm{sim}(H_{g,i}, H_{g,k})},
\end{equation}
where $B$ is the batch size,  $H_{\mathrm{mix},ij}$ is the representation of the mixed graph combining the $i$-th causal subgraph and the $j$-th confounding subgraph, and $H_{g,i}$ is the representation of the original graph.
% % The environmental consistency loss is defined as:
% \begin{equation}
%     L_{\text{env}} = \frac{1}{B} \sum_{i=1}^B \frac{1}{K(K-1)} \sum_{j=1}^K \sum_{l=1, l \neq j}^K \text{sim}(h_{\text{mix}, ij}, h_{\text{mix}, il}),
% \end{equation}
% where $h_{mix,ij}$ and $h_{mix,il}$ are representations of mixed graphs generated by pairing the $i$-th causal subgraph with the $j$-th and $l$-th confounding subgraphs, respectively.
\begin{remark}
The ultimate loss used in our paradigm is a simple combination of the GIB objective and the causal intervention loss:
$L = L_{\text{GIB}} + \lambda  L_{\text{CI}}$.
\end{remark}

% \subsection{Implementation details}

\begin{table*}[t]
\centering
% \caption{OOD generalization performance on GOOD-ZINC dataset. \textbf{Boldfaced values} denote the best performance, \underline{underlined values} indicate the second-best.}
\caption{OOD generalization performance on GOOD-ZINC dataset, with {\bf boldface} being the best and \underline{underline} being the runner-up.}
\label{good-zinc-results}
\begin{small}
\begin{sc}
\resizebox{\linewidth}{!}{%
\begin{tabular}{@{\hskip 0pt}lcccccccc@{\hskip 0pt}}
\toprule
\multirow{4}{*}{GOOD-ZINC} & \multicolumn{4}{c}{Scaffold} & \multicolumn{4}{c}{Size} \\
\cmidrule(lr){2-5} \cmidrule(lr){6-9}
 & \multicolumn{2}{c}{Covariate} & \multicolumn{2}{c}{Concept} & \multicolumn{2}{c}{Covariate} & \multicolumn{2}{c}{Concept} \\
\cmidrule(lr){2-3} \cmidrule(lr){4-5} \cmidrule(lr){6-7} \cmidrule(lr){8-9}
 & ID & OOD & ID & OOD & ID & OOD & ID & OOD \\ \midrule
ERM      & 0.1188±0.0030 & 0.1660±0.0093 & 0.1174±0.0013 & 0.1248±0.0018 & 0.1222±0.0061 & 0.2331±0.0169 & 0.1304±0.0010 & 0.1406±0.0002 \\
IRM      & 0.1258±0.0033 & 0.2313±0.0243 & 0.1176±0.0052 & 0.1245±0.0062 & 0.1217±0.0014 & 0.5840±0.0039 & 0.1331±0.0045 & 0.1338±0.0011 \\
VREx     & 0.0978±0.0016 & 0.1561±0.0021 & 0.1928±0.0021 & 0.1271±0.0020 & 0.1841±0.0009 & 0.2276±0.0005 & 0.1206±0.0008 & 0.1289±0.0039 \\
Mixup    & 0.1348±0.0025 & 0.2157±0.0098 & 0.1192±0.0026 & 0.1296±0.0049 & 0.1431±0.0070 & 0.2573±0.0042 & 0.1625±0.0121 & 0.1660±0.0063 \\
DANN     & 0.1152±0.0021 & 0.1734±0.0005 & 0.1284±0.0031 & 0.1289±0.0020 & 0.1053±0.0081 & 0.2254±0.0140 & 0.1227±0.0008 & 0.1271±0.0039 \\
Coral    & 0.1252±0.0043 & 0.1734±0.0034 & 0.1173±0.0029 & 0.1260±0.0024 & 0.1164±0.0004 & 0.2243±0.0147 & 0.1246±0.0062 & 0.1270±0.0020\\
CIGA     & 0.1568±0.0034 & 0.2986±0.0041 & 0.1926±0.0120 & 0.2415±0.0115 & 0.1500±0.0001 & 0.6102±0.0148 & 0.3560±0.0160 & 0.3240±0.0451 \\
DIR      & 0.2483±0.0056 & 0.3650±0.0032 & 0.2510±0.0001 & 0.2619±0.0076 & 0.2515±0.0529 & 0.4224±0.0679 & 0.4831±0.0823 & 0.3630±0.0872 \\
GSAT     & \underline{0.0890±0.0031} & \underline{0.1419±0.0043} & \underline{0.0928±0.0029} & \underline{0.0999±0.0029} & \underline{0.0876±0.0032} & \underline{0.2112±0.0033} & \underline{0.1002±0.0013} & \underline{0.1043±0.0001} \\ 
Ours     & \textbf{0.0514±0.0061} & \textbf{0.1046±0.0007} & \textbf{0.0659±0.0041} & \textbf{0.0518±0.0007} & \textbf{0.0466±0.0034} & \textbf{0.1484±0.0033} & \textbf{0.0577±0.0008} & \textbf{0.0580±0.0004} \\
\bottomrule
\end{tabular}%
}
\end{sc}
\end{small}
\end{table*}

\section{Experiments}
In this section, we evaluate the prediction performance and OOD generalization ability of our method. We comprehensively compare our method with existing models to demonstrate the superior generalization ability of our method on regression tasks. We briefly introduce the dataset, baselines, and experimental settings here.

\subsection{Datsets}
\textbf{GOOD-ZINC.}
GOOD-ZINC is a regression task in the GOOD benchmark ~\citep{gui2022good}, which aims to test the out-of-distribution performance of real-world molecular property regression datasets from the ZINC database ~\citep{gomez2018automatic}. The input is a molecular graph containing up to 38 heavy atoms, and the task is to predict the restricted solubility of the molecule ~\citep{jin2018junction,kusner2017grammar}. GOOD-ZINC includes four specific OOD types: Scaffold-Covariate, Scaffold-Concept, Size-Covariate, and Size-Concept. Scaffold OOD involves changes in molecular structures, while Size OOD varies graph size. Each can manifest as Covariate Shift ($P(X)$ changes, $P(Y | X)$ remains stable) or Concept Shift (spurious correlations in training break in testing).

\textbf{ReactionOOD-SOOD.}
In addition to the GOOD benchmark, we also used three S-OOD datasets in the ReactionOOD benchmark ~\citep{wang2023towards}, namely Cycloaddition ~\citep{stuyver2023reaction}, $\text{E2} \text{\&} \text{S}_{\text{N}}\text{2}$ ~\citep{von2020thousands}, and RDB7 ~\citep{spiekermann2022high}, which are designed to extract information outside the structural distribution during molecular reactions.  Cycloaddition and RDB7 have two domains: Total Atom Number (where the total number of atoms in a reaction exceeds the training range) and First Reactant Scaffold (where the first reactant has a new molecular scaffold unseen in training), while $\text{E2} \text{\&} \text{S}_{\text{N}}\text{2}$ dataset contains reactions with molecules whose scaffold cannot be properly defined, which prevents the scaffold from being an applicable domain index for this dataset. The definitions of two shifts Covariate and Concept in ReactionOOD are consistent with those in GOOD.

\subsection{Baselines and Setup}

As our framework is general and aims to address distribution shifts, we compare it against several baseline methods. Empirical Risk Minimization (ERM) ~\citep{vapnik1991principles} serves as a non-OOD baseline for comparison with OOD methods. We consider both Euclidean and graph-based state-of-the-art OOD approaches: (1) Euclidean OOD methods include IRM ~\citep{arjovsky2019invariant}, VREx ~\citep{krueger2021out}, GroupDRO ~\citep{sagawa2019distributionally}, DANN ~\citep{ganin2016domain}, Coral ~\citep{sun2016deep}, and Mixup ~\citep{zhang2017mixup}; (2) Graph OOD methods include CIGA ~\citep{chen2022learning}, GSAT ~\citep{miao2022interpretable}, and DIR ~\citep{wu2022discovering}.

For a fair comparison, all methods are implemented with consistent architectures and hyperparameters, ensuring that performance differences arise solely from the method itself. To provide reliable results, each experiment is repeated three times with different random seeds, and we report the mean and standard error of the results. Detailed settings and hyperparameter configurations are described in \Cref{appendix:settings}.

\begin{table*}[t]
\centering
\caption{OOD generalization performance on Cycloaddition and RDB7 dataset.}
\label{ood-results-main}
\begin{small}
\begin{sc}
\resizebox{\textwidth}{!}{%
\begin{tabular}{@{}llcccccccc@{}}
\toprule
\multirow{4}{*}{Dataset} & \multirow{4}{*}{Methods} & \multicolumn{4}{c}{First Reactant Scaffold} & \multicolumn{4}{c}{Total Atom Number} \\
\cmidrule(lr){3-6} \cmidrule(lr){7-10}
 & & \multicolumn{2}{c}{Covariate} & \multicolumn{2}{c}{Concept} & \multicolumn{2}{c}{Covariate} & \multicolumn{2}{c}{Concept} \\
\cmidrule(lr){3-4} \cmidrule(lr){5-6} \cmidrule(lr){7-8} \cmidrule(lr){9-10}
 & & ID & OOD & ID & OOD & ID & OOD & ID & OOD \\ \midrule
\multirow{10}{*}{Cycloaddition} 
& ERM      & \underline{4.38±0.04} & 4.80±0.38 & \underline{4.79±0.03} & \underline{5.60±0.02} & \textbf{3.77±0.01} & \textbf{4.36±0.15} & 4.22±0.04 & 5.69±0.03 \\
& IRM      & 15.30±0.05 & 21.16±0.01 & 17.55±0.03 & 18.64±0.25 & 17.53±0.17 & 17.44±0.14 & 23.14±0.02 & 22.56±0.01 \\
& VREx     & 5.54±0.02 & 6.69±0.48 & 5.02±0.05 & 6.14±0.09 & 4.79±0.03 & 5.22±0.06 & 4.92±0.14 & 6.39±0.04 \\
& Mixup    & 4.51±0.04 & 5.24±0.83 & 4.90±0.01 & 5.90±0.05 & 3.90±0.13 & 4.53±0.03 & \underline{4.11±0.09} & 5.93±0.13 \\
& DANN     & 4.42±0.03 & 4.68±0.12 & 4.81±0.01 & 5.75±0.06 & 3.87±0.05 & 4.65±0.10 & 4.18±0.02 & 5.68±0.10 \\
& Coral    & \textbf{4.36±0.07} & 4.95±0.30 & 4.82±0.03 & 5.72±0.16 & 4.39±0.59 & 5.05±0.48 & \textbf{4.10±0.05} & 5.74±0.04 \\
& CIGA     & 5.26±0.04 & 5.67±0.04 & 5.30±0.29 & 5.64±0.03 & 4.93±0.05 & 6.62±1.09 & 5.03±0.09 & 6.21±0.06 \\
& DIR      & 4.94±0.02 & 5.31±0.79 & 5.85±0.20 & 6.30±0.38 & 5.52±0.03 & 6.86±0.05 & 5.21±0.12 & 7.09±0.03 \\
& GSAT     & 4.42±0.05 & \underline{4.63±0.05} & 4.87±0.01 & 5.69±0.01 & \underline{3.81±0.01} & 4.56±0.01 & 4.12±0.04 & \underline{5.64±0.11} \\
& Ours     & 4.57±0.13 & \textbf{4.22±0.09} & \textbf{4.53±0.04} & \textbf{5.37±0.05} & 4.06±0.01 & \underline{4.42±0.24} & 4.41±0.22 & \textbf{5.53±0.12} \\ 
\midrule
\multirow{10}{*}{RDB7} 
& ERM      & 10.28±0.05 & 22.95±0.90 & 11.38±0.08 & \textbf{14.81±0.05} & 10.86±0.01 & \underline{7.66±0.55} & \textbf{11.28±0.15} & \underline{15.79±0.24} \\
& IRM      & 59.87±0.02 & 76.51±0.46 & 65.72±0.13 & 63.03±0.13 & 63.55±0.02 & 69.06±0.37 & 81.14±0.02 & 46.84±0.42 \\
& VREx     & 16.62±0.18 & \textbf{21.89±0.02} & 14.62±0.04 & 18.28±0.09 & 14.60±0.01 & 13.84±0.07 & 34.66±1.56 & 32.59±3.28 \\
& Mixup    & 10.76±0.07 & 23.49±0.09 & 11.89±0.05 & 15.64±0.10 & 11.13±0.02 & 10.78±0.17 & 	11.66±0.04 & 17.21±0.28 \\
& DANN     & \underline{10.28±0.05} & 23.54±0.07 & 11.28±0.01 & 14.93±0.05 & 10.77±0.22 & 8.29±0.10 & 11.34±0.05 & 16.28±0.15 \\
& Coral    & 10.30±0.12 & \underline{22.19±0.63} & \textbf{11.12±0.03} & 14.81±0.06 & \underline{10.61±0.01} & 8.04±0.14 & \underline{11.33±0.08} & 16.13±0.08 \\
& CIGA     & 14.97±0.75 & 30.08±0.84 & 18.68±1.94 & 21.35±1.34 & 16.48±0.69 & 19.12±1.85 & 20.58±1.54 & 18.53±1.30 \\
& DIR      & 14.34±0.68 & 26.99±0.49 & 17.13±1.76 & 20.18±1.86 & 14.03±2.06 & 15.01±0.98 & 13.52±0.51 & 16.60±1.09 \\
& GSAT     & 10.52±0.04 & 23.45±0.11 & 11.26±0.25 & \underline{14.85±0.12} & 10.80±0.01 & 8.66±0.10 & 11.58±0.03 & 16.08±0.41 \\
& Ours     & \textbf{10.12±0.08} & 23.11±0.46 & \underline{11.26±0.02} & 14.94±0.25 & \textbf{10.51±0.08} & \textbf{6.84±0.32} & 11.46±0.06 & \textbf{15.73±0.37} \\ 
\bottomrule
\end{tabular}
}
\end{sc}
\end{small}
\end{table*}

\begin{table}[t]
\centering
\caption{OOD generalization performance on $\text{E2} \text{\&} \text{S}_{\text{N}}\text{2}$ dataset.}
\label{ood-results-e2sn2}
\begin{small}
\begin{sc}
\resizebox{\columnwidth}{!}{%
\begin{tabular}{@{}lcccc@{}}
\toprule
\multirow{3}{*}{Methods} & \multicolumn{2}{c}{Covariate} & \multicolumn{2}{c}{Concept} \\
\cmidrule(lr){2-3} \cmidrule(lr){4-5}
 & ID & OOD & ID & OOD \\ \midrule
ERM      & 4.45±0.04 & 5.47±0.27 & 4.87±0.02 & 5.04±0.02 \\
IRM      & 11.61±0.18 & 21.54±1.07 & 20.95±0.02 & 17.57±0.03 \\
VREx     & 4.58±0.02 & 5.48±0.13 & 10.75±1.54 & 8.77±2.31 \\
Mixup    & 4.55±0.09 & 5.55±0.01 & 4.69±0.08 & 5.11±0.01 \\
DANN     & 4.51±0.06 & \underline{5.38±0.04} & \textbf{4.48±0.10} & 5.04±0.02 \\
Coral    & \underline{4.44±0.11} & 5.68±0.20 & 4.54±0.02 & \textbf{4.97±0.07} \\
CIGA     & 5.05±0.35 & 6.57±0.52 & 4.65±0.26 & 5.39±0.47 \\
DIR      & 5.61±0.26 & 6.59±0.31 & 6.56±0.34 & 6.29±0.11 \\
GSAT     & 4.55±0.01 & 5.69±0.05 & 4.55±0.09 & 5.04±0.03 \\
Ours     & \textbf{4.40±0.03} & \textbf{4.83±0.10} & \underline{4.53±0.12} & \underline{5.03±0.09} \\ 
\bottomrule
\end{tabular}
}
\end{sc}
\end{small}
\end{table}

% \begin{figure*}[h]
%     \centering
%     \includegraphics[trim=0 0 0 0,clip,width=0.75\textwidth]{fig/classification_ib_results.pdf}
%     \caption{Ablation study on confounder predictive power (left) and causal intervention methods (right) for OOD generalization on GOOD-Motif.}
%     \label{fig:ablation}
% \end{figure*}

\subsection{Results of GOOD}
\label{sec:exp-good}

As shown in \Cref{good-zinc-results}, our proposed method achieves SOTA performance on GOOD-ZINC, consistently outperforming all baseline methods across both domains (Scaffold and Size) and under different distribution shifts (Covariate and Concept). Specifically, in terms of Mean Absolute Error (MAE), our method demonstrates significant improvements in both in-distribution (ID) and out-of-distribution (OOD) settings.

For instance, in the Scaffold domain under the Covariate shift, our method achieves an MAE of 0.0514±0.0061 (ID) and 0.1046±0.0007 (OOD), outperforming GSAT, the next-best method, by 42.2\% in ID and 26.3\% in OOD performance. Similarly, under the Concept shift, our method achieves 0.0659±0.0041 (ID) and 0.0518±0.0007 (OOD), representing improvements of 29.0\% and 48.1\%, respectively, over GSAT.

In the Size domain, our method also achieves remarkable results. Under the Covariate shift, it achieves an MAE of 0.0466±0.0034 (ID) and 0.1484±0.0033 (OOD), which translate to 46.8\% lower ID error and 29.7\% lower OOD error compared to GSAT. Similarly, under the Concept shift, our approach yields an MAE of 0.0577±0.0008 (ID) and 0.0580±0.0004 (OOD), improving upon GSAT by 42.4\% and 44.4\%, respectively.

In addition to achieving lower MAE values, our method exhibits significantly reduced variances compared to other approaches, highlighting its stability under diverse conditions. These findings confirm the strong generalization capability of our method across different domains and types of distributional shifts.

\subsection{Results of ReactionOOD}
\label{sec:exp-rood}

% Tables \ref{ood-results-main} and \ref{ood-results-e2sn2} demonstrate the generalization ability of our method across various datasets and evaluation settings, as measured by RMSE. In the Cycloaddition dataset, under the Size domain with a Concept shift, OURS achieves an OOD RMSE of 5.53 ± 0.12, outperforming all competing methods, despite its ID RMSE (4.41 ± 0.22) being marginally lower than the best-performing models. Similarly, in the RDB7 dataset under the Concept shift, OURS achieves the lowest OOD RMSE (15.73 ± 0.37), highlighting its effectiveness in addressing distributional discrepancies, even though its ID performance does not surpass ERM. For the E2SN2 dataset, OURS delivers the best OOD RMSE (4.83 ± 0.10) under the Covariate shift and competitive results under the Concept shift (5.03 ± 0.09), while maintaining stable ID performance (4.40 ± 0.03 for Covariate and 4.53 ± 0.12 for Concept).

% It is important to note that our method does not achieve the absolute best performance in a few specific OOD scenarios. This can likely be attributed to the inherent complexity and variability of OOD tasks, where certain methods may be better suited to particular types of spurious correlations or dataset-specific characteristics. Nonetheless, our method distinguishes itself through its overall stability and reliability, delivering robust and consistent results across a wide range of scenarios without significant trade-offs between ID and OOD performance.

\Cref{ood-results-main} and \Cref{ood-results-e2sn2} highlight the robust generalization ability of our method across multiple datasets and evaluation settings, as measured by RMSE. Our method achieves the best OOD performance in 6 out of 10 cases and ranks second in 2 cases. Notably, in cases where another method outperforms ours, the performance gap is within a small margin.

For instance, in the Cycloaddition dataset, under the total atom number domain with a concept shift, Our method achieves an OOD RMSE of 5.53 ± 0.12, outperforming all baseline methods. While some non-causal baselines (e.g., Coral in this specific setting, achieving an ID RMSE of 4.10 ± 0.05 versus our 4.41 ± 0.22) might get better ID performance by exploiting spurious but predictive features, such approaches can become less reliable under OOD conditions (e.g., Coral's OOD RMSE degrades to 5.74 ± 0.04). In contrast, our method's focus on identifying and removing these spurious features contributes to its stable and superior OOD performance. Even in other Cycloaddition cases where ours ranks second, such as the same domain with a covariate shift, the OOD RMSE (4.42 ± 0.24) is only 0.06 away from the best-performing method (4.36 ± 0.15).

% Even in cases where Ours ranks second, such as the same domain with a covariate shift, the RMSE (4.42 ± 0.24) is only 0.06 away from the best-performing method (4.36 ± 0.15).

In RDB7, a smaller dataset within the ReactionOOD where causal inference can be more difficult, our method achieves the lowest OOD RMSE (15.73 ± 0.37) under the concept shift. Our method's principled focus on true causal features, which leads to better OOD generalization ability and stability. Even though causal methods generally face challenges in smaller datasets~\citep{guo2020survey}, our approach consistently outperforms other listed causal intervention baselines such as CIGA in all RDB7 settings. In the $\text{E2} \text{\&} \text{S}_{\text{N}}\text{2}$ dataset, our method delivers the best OOD RMSE (4.83 ± 0.10) under the covariate shift and achieves highly competitive results under the concept shift (5.03 ± 0.09).

% Similarly, in the RDB7 dataset, our method achieves the lowest OOD RMSE (15.73 ± 0.37) under the concept shift, outperforming other methods, even though its ID performance is slightly behind ERM. In the E2SN2 dataset, our method delivers the best OOD RMSE (4.83 ± 0.10) under the covariate shift and achieves highly competitive results under the concept shift (5.03 ± 0.09). Across all datasets, our method maintains stable ID performance, demonstrating that its strong OOD generalization does not come at the cost of ID.

% Overall, our method distinguishes itself through consistently strong performance across various settings, achieving either the best or highly competitive results in nearly all cases. This highlights the robustness and reliability of our method, making it a well-rounded and practical solution for addressing OOD challenges without significant trade-offs.

As noted in OOD-GNN~\citep{tajwar2021no}, no method consistently performs best on every dataset due to varying distribution shifts and inductive biases. Our approach, designed under more general and weaker assumptions which do not assume that spurious features are non-predictive, aims to tackle a wider range of real-world distribution shifts.
% \begin{figure*}[h]
%     \centering
%     \includegraphics[trim=0 0 0 0,clip,width=0.75\textwidth]{fig/classification_ib_results.pdf}
%     \caption{Ablation study on confounder predictive power (left) and causal intervention methods (right) for OOD generalization on GOOD-Motif.}
%     \label{fig:ablation}
% \end{figure*}

\begin{figure*}[htbp]
    \centering
    \includegraphics[width=0.4\textwidth]{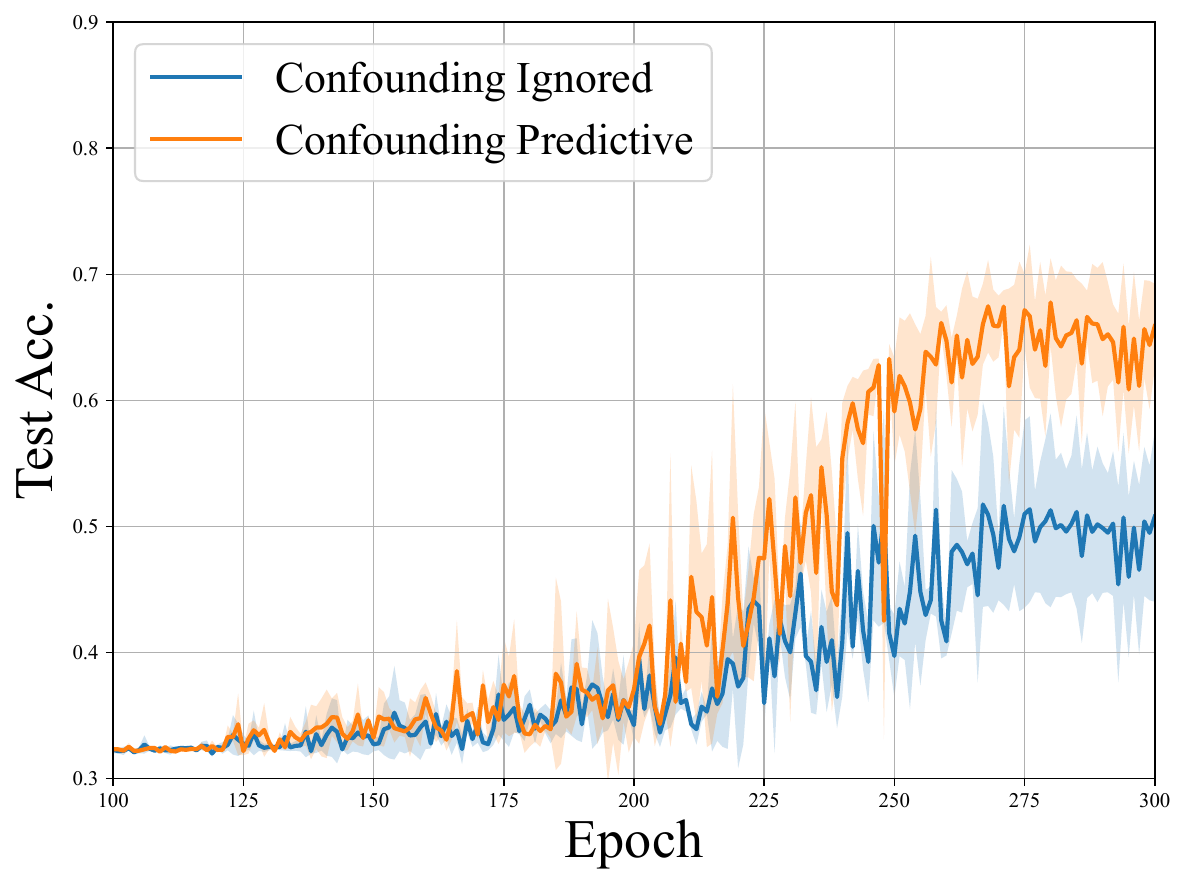}
    \hspace{0.05\textwidth}
    \includegraphics[width=0.4\textwidth]{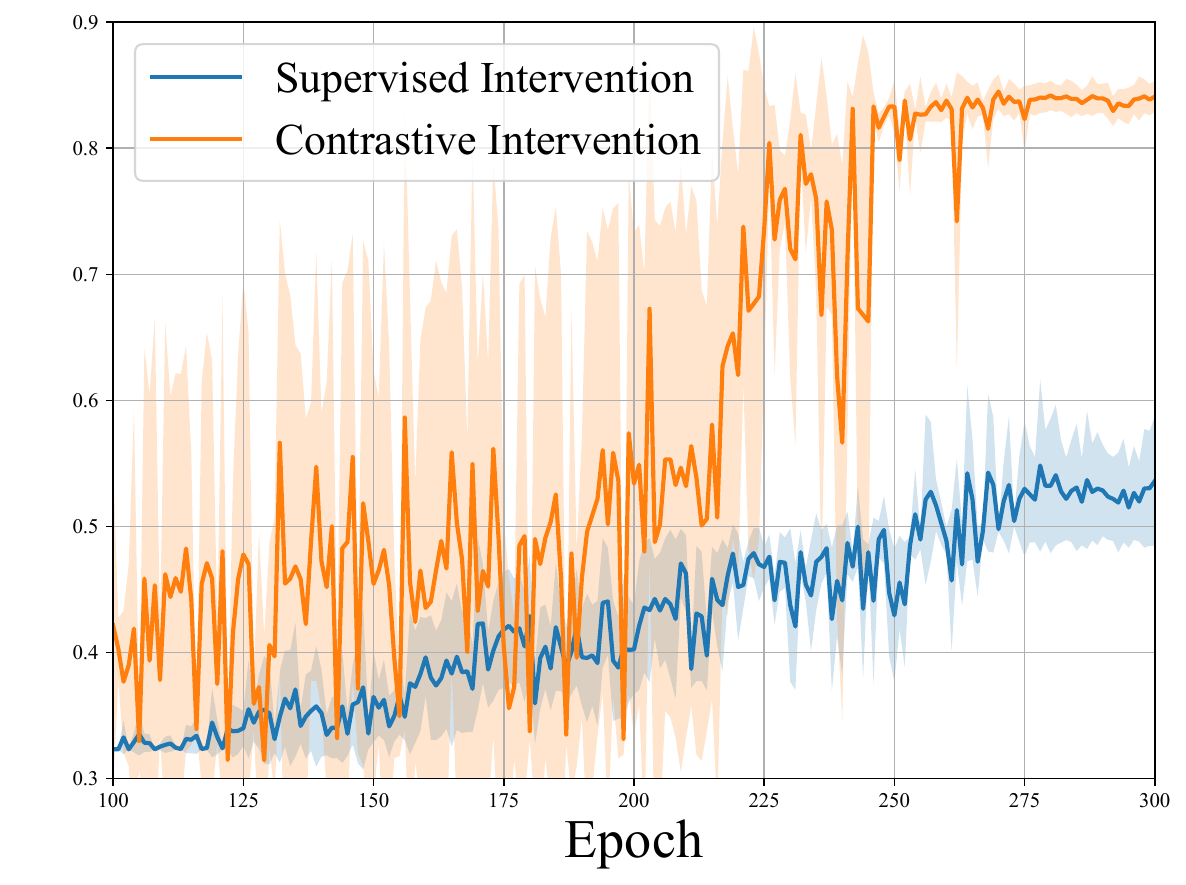}
    \caption{
        Ablation study on confounder predictive power (left) and causal intervention methods (right) for OOD generalization on GOOD-Motif.
    }
    \label{fig:ablation}
\end{figure*}

\subsection{Effectiveness of OURS in Classification Task}

To validate the generality and effectiveness of our proposed losses, $L_{\mathrm{GIB}}$ and $L_{\mathrm{CI}}$, we conduct ablation studies on the GOOD-Motif dataset under the size domain setting. The results, evaluated in terms of accuracy, are reported on the OOD dataset, as shown in \Cref{fig:ablation}. The ablation study on $L_{\mathrm{GIB}}$ aims to examine our hypothesis that confounders possess certain predictive power; thus, this experiment excludes the causal intervention loss $L_{\mathrm{CI}}$ Conversely, the ablation study on $L_{\mathrm{CI}}$ evaluates whether the contrastive learning-driven causal intervention loss can independently achieve strong OOD performance. Therefore, in this experiment, we do not incorporate the predictive power of confounding factors.

% To validate the effectiveness of our method\tyq{be specific, which part}, we conducted two ablation studies on classification tasks, as shown in Fig. \ref{fig:ablation}. 

\paragraph{Predictive power of confounding subgraphs.}
The left panel compares minimizing confounding subgraph prediction alone versus introducing constraints to model their predictive ability. The results show that ignore the predictive role of confounding subgraphs leads to incomplete disentanglement and weaker OOD generalization, demonstrating that accounting for their influence is crucial.

\paragraph{Effectiveness of contrastive learning.}
The right panel compares using predictions from randomly generated counterfactual graphs as causal intervention loss versus our proposed contrastive learning loss. The results show that our contrastive learning approach, initially validated in regression tasks, is equally effective in classification tasks, highlighting its general applicability.

These studies confirm the importance of explicitly modeling confounding subgraphs and the robustness of our contrastive learning loss for OOD generalization. More experimental results are provided in the \Cref{appendix:ablation}.

\section{Conclusion}

In this work, we propose a recipe for causal graph regression through reshaping the processing of confounding effects in existing CGL classification-specific techniques.
In particular, we develop an enhanced graph information bottleneck (GIB) loss function which highlights the impact of confounding effects and consequently benefits the recognition of causal subgraphs.
% on out-of-distribution (OOD) generalization.
Moreover, we revisit the causal intervention technique, which randomly combines causal subgraphs and confounder from the same class (label) to eliminate confounding effects.
Adapting this technique to regression requires removal of label information; 
to this end, we analyze the principle of causal intervention and propose to connect it with unsupervised contrastive learning loss. 
% To address OOD problem on graph regression tasks, we propose a novel approach that incorporates an enhanced Graph Information Bottleneck (GIB) objective to harness the predictive power of confounders while mitigating their adverse effects. Additionally, we introduce a causal intervention loss, guided by contrastive learning, to dynamically focus on causal subgraphs during regression. 
% The two approaches together form our proposed
Experimental results on graph OOD benchmarks demonstrate the effectiveness of our proposed techniques in improving the generalizability of graph regression models.
% our ablation studies further confirm that the proposals can 
% Future work includes exploring more advanced causal inference techniques and extending our approach to other graph-based tasks, such as node regression.

\section*{Acknowledgements}
\label{section-ack}

We sincerely thank the area chair and the anonymous reviewers for their valuable feedback and constructive suggestions, which helped improve this work.
The authors acknowledge funding from Research Grants Council (RGC) under grant \texttt{22303424} and GuangDong Basic and Applied Basic Research Foundation under grant \texttt{2025A1515010259}.

% \newpage
\section*{Impact Statement}

% Sample:

This paper presents work whose goal is to advance the field of 
Machine Learning. There are many potential societal consequences 
of our work, none which we feel must be specifically highlighted here.


\bibliographystyle{icml2025}
\bibliography{references}

\begin{thebibliography}{53}
\providecommand{\natexlab}[1]{#1}
\providecommand{\url}[1]{\texttt{#1}}
\expandafter\ifx\csname urlstyle\endcsname\relax
  \providecommand{\doi}[1]{doi: #1}\else
  \providecommand{\doi}{doi: \begingroup \urlstyle{rm}\Url}\fi

\bibitem[Amini et~al.(2020)Amini, Schwarting, Soleimany, and Rus]{amini2020deep}
Amini, A., Schwarting, W., Soleimany, A., and Rus, D.
\newblock Deep evidential regression.
\newblock \emph{Advances in neural information processing systems}, 33:\penalty0 14927--14937, 2020.

\bibitem[Arjovsky et~al.(2019)Arjovsky, Bottou, Gulrajani, and Lopez-Paz]{arjovsky2019invariant}
Arjovsky, M., Bottou, L., Gulrajani, I., and Lopez-Paz, D.
\newblock Invariant risk minimization.
\newblock \emph{arXiv preprint arXiv:1907.02893}, 2019.

\bibitem[Chechik et~al.(2003)Chechik, Globerson, Tishby, and Weiss]{chechik2003information}
Chechik, G., Globerson, A., Tishby, N., and Weiss, Y.
\newblock Information bottleneck for gaussian variables.
\newblock \emph{Advances in Neural Information Processing Systems}, 16, 2003.

\bibitem[Chen et~al.(2022)Chen, Zhang, Bian, Yang, Kaili, Xie, Liu, Han, and Cheng]{chen2022learning}
Chen, Y., Zhang, Y., Bian, Y., Yang, H., Kaili, M., Xie, B., Liu, T., Han, B., and Cheng, J.
\newblock Learning causally invariant representations for out-of-distribution generalization on graphs.
\newblock \emph{Advances in Neural Information Processing Systems}, 35:\penalty0 22131--22148, 2022.

\bibitem[Chen et~al.(2024)Chen, Bian, Zhou, Xie, Han, and Cheng]{chen2024does}
Chen, Y., Bian, Y., Zhou, K., Xie, B., Han, B., and Cheng, J.
\newblock Does invariant graph learning via environment augmentation learn invariance?
\newblock \emph{Advances in Neural Information Processing Systems}, 36, 2024.

\bibitem[Fan et~al.(2022)Fan, Wang, Mo, Shi, and Tang]{fan2022debiasing}
Fan, S., Wang, X., Mo, Y., Shi, C., and Tang, J.
\newblock Debiasing graph neural networks via learning disentangled causal substructure.
\newblock \emph{Advances in Neural Information Processing Systems}, 35:\penalty0 24934--24946, 2022.

\bibitem[Fan et~al.(2023)Fan, Wang, Shi, Cui, and Wang]{fan2023generalizing}
Fan, S., Wang, X., Shi, C., Cui, P., and Wang, B.
\newblock Generalizing graph neural networks on out-of-distribution graphs.
\newblock \emph{IEEE Transactions on Pattern Analysis and Machine Intelligence}, 2023.

\bibitem[Ganin et~al.(2016)Ganin, Ustinova, Ajakan, Germain, Larochelle, Laviolette, March, and Lempitsky]{ganin2016domain}
Ganin, Y., Ustinova, E., Ajakan, H., Germain, P., Larochelle, H., Laviolette, F., March, M., and Lempitsky, V.
\newblock Domain-adversarial training of neural networks.
\newblock \emph{Journal of machine learning research}, 17\penalty0 (59):\penalty0 1--35, 2016.

\bibitem[Gao et~al.(2023)Gao, Li, Qiang, Si, Xu, Zheng, and Sun]{gao2023robust}
Gao, H., Li, J., Qiang, W., Si, L., Xu, B., Zheng, C., and Sun, F.
\newblock Robust causal graph representation learning against confounding effects.
\newblock In \emph{Proceedings of the AAAI Conference on Artificial Intelligence}, volume~37, pp.\  7624--7632, 2023.

\bibitem[G{\'o}mez-Bombarelli et~al.(2018)G{\'o}mez-Bombarelli, Wei, Duvenaud, Hern{\'a}ndez-Lobato, S{\'a}nchez-Lengeling, Sheberla, Aguilera-Iparraguirre, Hirzel, Adams, and Aspuru-Guzik]{gomez2018automatic}
G{\'o}mez-Bombarelli, R., Wei, J.~N., Duvenaud, D., Hern{\'a}ndez-Lobato, J.~M., S{\'a}nchez-Lengeling, B., Sheberla, D., Aguilera-Iparraguirre, J., Hirzel, T.~D., Adams, R.~P., and Aspuru-Guzik, A.
\newblock Automatic chemical design using a data-driven continuous representation of molecules.
\newblock \emph{ACS central science}, 4\penalty0 (2):\penalty0 268--276, 2018.

\bibitem[Gui et~al.(2022)Gui, Li, Wang, and Ji]{gui2022good}
Gui, S., Li, X., Wang, L., and Ji, S.
\newblock Good: A graph out-of-distribution benchmark.
\newblock \emph{Advances in Neural Information Processing Systems}, 35:\penalty0 2059--2073, 2022.

\bibitem[Guo et~al.(2020)Guo, Cheng, Li, Hahn, and Liu]{guo2020survey}
Guo, R., Cheng, L., Li, J., Hahn, P.~R., and Liu, H.
\newblock A survey of learning causality with data: Problems and methods.
\newblock \emph{ACM Computing Surveys (CSUR)}, 53\penalty0 (4):\penalty0 1--37, 2020.

\bibitem[Guo et~al.(2025)Guo, Wu, Xiao, Aggarwal, Liu, and Wang]{guo2025counterfactual}
Guo, Z., Wu, Z., Xiao, T., Aggarwal, C., Liu, H., and Wang, S.
\newblock Counterfactual learning on graphs: A survey.
\newblock \emph{Machine Intelligence Research}, 22\penalty0 (1):\penalty0 17--59, 2025.

\bibitem[Jin et~al.(2018)Jin, Barzilay, and Jaakkola]{jin2018junction}
Jin, W., Barzilay, R., and Jaakkola, T.
\newblock Junction tree variational autoencoder for molecular graph generation.
\newblock In \emph{International conference on machine learning}, pp.\  2323--2332. PMLR, 2018.

\bibitem[Kingma et~al.(2013)Kingma, Welling, et~al.]{kingma2013auto}
Kingma, D.~P., Welling, M., et~al.
\newblock Auto-encoding variational bayes, 2013.

\bibitem[Krueger et~al.(2021)Krueger, Caballero, Jacobsen, Zhang, Binas, Zhang, Le~Priol, and Courville]{krueger2021out}
Krueger, D., Caballero, E., Jacobsen, J.-H., Zhang, A., Binas, J., Zhang, D., Le~Priol, R., and Courville, A.
\newblock Out-of-distribution generalization via risk extrapolation (rex).
\newblock In \emph{International conference on machine learning}, pp.\  5815--5826. PMLR, 2021.

\bibitem[Kusner et~al.(2017)Kusner, Paige, and Hern{\'a}ndez-Lobato]{kusner2017grammar}
Kusner, M.~J., Paige, B., and Hern{\'a}ndez-Lobato, J.~M.
\newblock Grammar variational autoencoder.
\newblock In \emph{International conference on machine learning}, pp.\  1945--1954. PMLR, 2017.

\bibitem[Li et~al.(2021)Li, Knoop, and Van~Lint]{li2021multistep}
Li, G., Knoop, V.~L., and Van~Lint, H.
\newblock Multistep traffic forecasting by dynamic graph convolution: Interpretations of real-time spatial correlations.
\newblock \emph{Transportation Research Part C: Emerging Technologies}, 128:\penalty0 103185, 2021.

\bibitem[Li et~al.(2022)Li, Wang, Zhang, and Zhu]{li2022ood}
Li, H., Wang, X., Zhang, Z., and Zhu, W.
\newblock Ood-gnn: Out-of-distribution generalized graph neural network.
\newblock \emph{IEEE Transactions on Knowledge and Data Engineering}, 35\penalty0 (7):\penalty0 7328--7340, 2022.

\bibitem[Lin et~al.(2021)Lin, Lan, and Li]{lin2021generative}
Lin, W., Lan, H., and Li, B.
\newblock Generative causal explanations for graph neural networks.
\newblock In \emph{International Conference on Machine Learning}, pp.\  6666--6679. PMLR, 2021.

\bibitem[Luo et~al.(2020)Luo, Cheng, Xu, Yu, Zong, Chen, and Zhang]{luo2020parameterized}
Luo, D., Cheng, W., Xu, D., Yu, W., Zong, B., Chen, H., and Zhang, X.
\newblock Parameterized explainer for graph neural network.
\newblock \emph{Advances in neural information processing systems}, 33:\penalty0 19620--19631, 2020.

\bibitem[Ma et~al.(2024)Ma, Li, and Ilyas]{ma2024utilizing}
Ma, F., Li, H., and Ilyas, M.
\newblock Utilizing reinforcement learning and causal graph networks to address the intricate dynamics in financial risk prediction.
\newblock \emph{International Journal of Information Technologies and Systems Approach (IJITSA)}, 17\penalty0 (1):\penalty0 1--19, 2024.

\bibitem[Ma(2024)]{ma2024survey}
Ma, J.
\newblock A survey of out-of-distribution generalization for graph machine learning from a causal view.
\newblock \emph{arXiv preprint arXiv:2409.09858}, 2024.

\bibitem[Miao et~al.(2022)Miao, Liu, and Li]{miao2022interpretable}
Miao, S., Liu, M., and Li, P.
\newblock Interpretable and generalizable graph learning via stochastic attention mechanism.
\newblock In \emph{International Conference on Machine Learning}, pp.\  15524--15543. PMLR, 2022.

\bibitem[Mitrovic et~al.(2020)Mitrovic, McWilliams, Walker, Buesing, and Blundell]{mitrovic2020representation}
Mitrovic, J., McWilliams, B., Walker, J., Buesing, L., and Blundell, C.
\newblock Representation learning via invariant causal mechanisms.
\newblock \emph{arXiv preprint arXiv:2010.07922}, 2020.

\bibitem[Nix \& Weigend(1994)Nix and Weigend]{nix1994learning}
Nix, D. and Weigend, A.
\newblock Learning local error bars for nonlinear regression.
\newblock \emph{Advances in neural information processing systems}, 7, 1994.

\bibitem[Oord et~al.(2018)Oord, Li, and Vinyals]{oord2018representation}
Oord, A. v.~d., Li, Y., and Vinyals, O.
\newblock Representation learning with contrastive predictive coding.
\newblock \emph{arXiv preprint arXiv:1807.03748}, 2018.

\bibitem[Pearl(2014)]{pearl2014interpretation}
Pearl, J.
\newblock Interpretation and identification of causal mediation.
\newblock \emph{Psychological methods}, 19\penalty0 (4):\penalty0 459, 2014.

\bibitem[Pleiss et~al.(2019)Pleiss, Souza, Kim, Li, and Weinberger]{pleiss2019neural}
Pleiss, G., Souza, A., Kim, J., Li, B., and Weinberger, K.~Q.
\newblock Neural network out-of-distribution detection for regression tasks.
\newblock 2019.

\bibitem[Qiao et~al.(2024)Qiao, Wang, and Li]{qiao2024causal}
Qiao, G., Wang, G., and Li, Y.
\newblock Causal enhanced drug--target interaction prediction based on graph generation and multi-source information fusion.
\newblock \emph{Bioinformatics}, 40\penalty0 (10):\penalty0 btae570, 2024.

\bibitem[Rollins et~al.(2024)Rollins, Cheng, and Metwally]{rollins2024molprop}
Rollins, Z.~A., Cheng, A.~C., and Metwally, E.
\newblock Molprop: Molecular property prediction with multimodal language and graph fusion.
\newblock \emph{Journal of Cheminformatics}, 16\penalty0 (1):\penalty0 56, 2024.

\bibitem[Rosenblatt(1958)]{rosenblatt1958perceptron}
Rosenblatt, F.
\newblock The perceptron: a probabilistic model for information storage and organization in the brain.
\newblock \emph{Psychological review}, 65\penalty0 (6):\penalty0 386, 1958.

\bibitem[Sagawa et~al.(2019)Sagawa, Koh, Hashimoto, and Liang]{sagawa2019distributionally}
Sagawa, S., Koh, P.~W., Hashimoto, T.~B., and Liang, P.
\newblock Distributionally robust neural networks for group shifts: On the importance of regularization for worst-case generalization.
\newblock \emph{arXiv preprint arXiv:1911.08731}, 2019.

\bibitem[Spiekermann et~al.(2022)Spiekermann, Pattanaik, and Green]{spiekermann2022high}
Spiekermann, K., Pattanaik, L., and Green, W.~H.
\newblock High accuracy barrier heights, enthalpies, and rate coefficients for chemical reactions.
\newblock \emph{Scientific Data}, 9\penalty0 (1):\penalty0 417, 2022.

\bibitem[Stuyver et~al.(2023)Stuyver, Jorner, and Coley]{stuyver2023reaction}
Stuyver, T., Jorner, K., and Coley, C.~W.
\newblock Reaction profiles for quantum chemistry-computed [3+ 2] cycloaddition reactions.
\newblock \emph{Scientific Data}, 10\penalty0 (1):\penalty0 66, 2023.

\bibitem[Sui et~al.(2022)Sui, Wang, Wu, Lin, He, and Chua]{sui2022causal}
Sui, Y., Wang, X., Wu, J., Lin, M., He, X., and Chua, T.-S.
\newblock Causal attention for interpretable and generalizable graph classification.
\newblock In \emph{Proceedings of the 28th ACM SIGKDD Conference on Knowledge Discovery and Data Mining}, pp.\  1696--1705, 2022.

\bibitem[Sui et~al.(2024)Sui, Mao, Wang, Wang, Wu, He, and Chua]{sui2024enhancing}
Sui, Y., Mao, W., Wang, S., Wang, X., Wu, J., He, X., and Chua, T.-S.
\newblock Enhancing out-of-distribution generalization on graphs via causal attention learning.
\newblock \emph{ACM Transactions on Knowledge Discovery from Data}, 18\penalty0 (5):\penalty0 1--24, 2024.

\bibitem[Sun \& Saenko(2016)Sun and Saenko]{sun2016deep}
Sun, B. and Saenko, K.
\newblock Deep coral: Correlation alignment for deep domain adaptation.
\newblock In \emph{Computer Vision--ECCV 2016 Workshops: Amsterdam, The Netherlands, October 8-10 and 15-16, 2016, Proceedings, Part III 14}, pp.\  443--450. Springer, 2016.

\bibitem[Tajwar et~al.(2021)Tajwar, Kumar, Xie, and Liang]{tajwar2021no}
Tajwar, F., Kumar, A., Xie, S.~M., and Liang, P.
\newblock No true state-of-the-art? ood detection methods are inconsistent across datasets.
\newblock \emph{arXiv preprint arXiv:2109.05554}, 2021.

\bibitem[Tishby \& Zaslavsky(2015)Tishby and Zaslavsky]{tishby2015deep}
Tishby, N. and Zaslavsky, N.
\newblock Deep learning and the information bottleneck principle.
\newblock In \emph{2015 ieee information theory workshop (itw)}, pp.\  1--5. IEEE, 2015.

\bibitem[Tishby et~al.(2000)Tishby, Pereira, and Bialek]{tishby2000information}
Tishby, N., Pereira, F.~C., and Bialek, W.
\newblock The information bottleneck method.
\newblock \emph{arXiv preprint physics/0004057}, 2000.

\bibitem[Vapnik(1991)]{vapnik1991principles}
Vapnik, V.
\newblock Principles of risk minimization for learning theory.
\newblock \emph{Advances in neural information processing systems}, 4, 1991.

\bibitem[von Rudorff et~al.(2020)von Rudorff, Heinen, Bragato, and von Lilienfeld]{von2020thousands}
von Rudorff, G.~F., Heinen, S.~N., Bragato, M., and von Lilienfeld, O.~A.
\newblock Thousands of reactants and transition states for competing e2 and s2 reactions.
\newblock \emph{Machine Learning: Science and Technology}, 1\penalty0 (4):\penalty0 045026, 2020.

\bibitem[Wang \& Veitch(2022)Wang and Veitch]{wang2022unified}
Wang, Z. and Veitch, V.
\newblock A unified causal view of domain invariant representation learning.
\newblock 2022.

\bibitem[Wang et~al.(2023)Wang, Chen, Duan, Li, Han, Cheng, and Tong]{wang2023towards}
Wang, Z., Chen, Y., Duan, Y., Li, W., Han, B., Cheng, J., and Tong, H.
\newblock Towards out-of-distribution generalizable predictions of chemical kinetics properties.
\newblock \emph{arXiv preprint arXiv:2310.03152}, 2023.

\bibitem[Wu et~al.(2022{\natexlab{a}})Wu, Zhang, Yan, and Wipf]{wu2022handling}
Wu, Q., Zhang, H., Yan, J., and Wipf, D.
\newblock Handling distribution shifts on graphs: An invariance perspective.
\newblock \emph{arXiv preprint arXiv:2202.02466}, 2022{\natexlab{a}}.

\bibitem[Wu et~al.(2020)Wu, Ren, Li, and Leskovec]{wu2020graph}
Wu, T., Ren, H., Li, P., and Leskovec, J.
\newblock Graph information bottleneck.
\newblock \emph{Advances in Neural Information Processing Systems}, 33:\penalty0 20437--20448, 2020.

\bibitem[Wu et~al.(2022{\natexlab{b}})Wu, Wang, Zhang, He, and Chua]{wu2022discovering}
Wu, Y.-X., Wang, X., Zhang, A., He, X., and Chua, T.-S.
\newblock Discovering invariant rationales for graph neural networks.
\newblock \emph{arXiv preprint arXiv:2201.12872}, 2022{\natexlab{b}}.

\bibitem[Wu et~al.(2022{\natexlab{c}})Wu, Wang, Zhang, Hu, Feng, He, and Chua]{wu2022deconfounding}
Wu, Y.-X., Wang, X., Zhang, A., Hu, X., Feng, F., He, X., and Chua, T.-S.
\newblock Deconfounding to explanation evaluation in graph neural networks.
\newblock \emph{arXiv preprint arXiv:2201.08802}, 2022{\natexlab{c}}.

\bibitem[Yu et~al.(2024)Yu, Yu, L{\o}kse, Jenssen, and Pr{\'\i}ncipe]{yu2024cauchy}
Yu, S., Yu, X., L{\o}kse, S., Jenssen, R., and Pr{\'\i}ncipe, J.~C.
\newblock Cauchy-schwarz divergence information bottleneck for regression.
\newblock In \emph{ICLR}, 2024.

\bibitem[Zhang(2017)]{zhang2017mixup}
Zhang, H.
\newblock mixup: Beyond empirical risk minimization.
\newblock \emph{arXiv preprint arXiv:1710.09412}, 2017.

\bibitem[Zhang et~al.(2023)Zhang, Chen, Mei, Luo, and Wei]{zhang2023regexplainer}
Zhang, J., Chen, Z., Mei, H., Luo, D., and Wei, H.
\newblock Regexplainer: Generating explanations for graph neural networks in regression task.
\newblock \emph{arXiv preprint arXiv:2307.07840}, 2023.

\bibitem[Zhao et~al.(2024)Zhao, Prapas, Karasante, Xiong, Papoutsis, Camps-Valls, and Zhu]{zhao2024causal}
Zhao, S., Prapas, I., Karasante, I., Xiong, Z., Papoutsis, I., Camps-Valls, G., and Zhu, X.~X.
\newblock Causal graph neural networks for wildfire danger prediction.
\newblock \emph{arXiv preprint arXiv:2403.08414}, 2024.

\end{thebibliography}

%%%%%%%%%%%%%%%%%%%%%%%%%%%%%%%%%%%%%%%%%%%%%%%%%%%%%%%%%%%%%%%%%%%%%%%%%%%%%%%
%%%%%%%%%%%%%%%%%%%%%%%%%%%%%%%%%%%%%%%%%%%%%%%%%%%%%%%%%%%%%%%%%%%%%%%%%%%%%%%
% APPENDIX
%%%%%%%%%%%%%%%%%%%%%%%%%%%%%%%%%%%%%%%%%%%%%%%%%%%%%%%%%%%%%%%%%%%%%%%%%%%%%%%
%%%%%%%%%%%%%%%%%%%%%%%%%%%%%%%%%%%%%%%%%%%%%%%%%%%%%%%%%%%%%%%%%%%%%%%%%%%%%%%
\newpage
\appendix
\onecolumn

\section{Supplementary Experiments}
\subsection{GOOD Benchmark}
The Graph Out-Of-Distribution (GOOD) benchmark is the most comprehensive and authoritative benchmark for assessing the OOD generalization of graph learning models. It includes 11 datasets, covering six graph-level and five node-level tasks, with 51 dataset splits across covariate shift, concept shift, and no shift scenarios. Among them, nine datasets focus on classification (binary and multi-class), one (GOOD-ZINC) on regression, and one (GOOD-PCBA) on multi-objective binary classification. GOOD is the first benchmark to incorporate both covariate and concept shifts within the same domain, enabling controlled comparisons. It evaluates 10 state-of-the-art OOD methods, including four tailored for graphs, resulting in 510 dataset-model combinations. As a result, GOOD provides a systematic and rigorous framework for benchmarking OOD generalization in graph learning

\subsection{ReactionOOD Benchmark}
The ReactionOOD benchmark is a specialized out-of-distribution (OOD) evaluation framework designed to systematically assess the generalization capabilities of machine learning models in predicting the kinetic properties of chemical reactions. It introduces three distinct levels of OOD shifts—structural, conditional, and mechanistic—and comprises six datasets, all formulated as regression tasks. Structural OOD (S-OOD) examines variations in reactant structures, including shifts based on total atomic count (E2 \& SN2) and reactant scaffolds (RDB7, Cycloaddition). Conditional OOD (C-OOD) investigates the effect of environmental conditions on kinetic properties, considering shifts in temperature (RMG Lib. T) and combined temperature-pressure settings (RMG Lib. TP). Mechanistic OOD (M-OOD) explores the impact of different reaction mechanisms (RMG Family) on kinetic property predictions.

\subsection{GOOD-ZINC Dataset Details}
\Cref{fig:zinc details} presents the number of graphs/nodes in different dataset splits for the GOOD-ZINC dataset. The dataset is analyzed under three types of distribution shifts: covariate, concept, and no shift. Each row represents the number of graphs/nodes in training, in-distribution (ID) validation, ID test, out-of-distribution (OOD) validation, and OOD test sets. The no-shift scenario serves as a baseline with no distributional difference between training and test sets.
\begin{table}[h]
    \centering
    \caption{Details of GOOD-ZINC dataset.}
    \begin{tabular}{llccccc}
        \hline
        \textbf{Dataset} & \textbf{Shift} & \textbf{Train} & \textbf{ID validation} & \textbf{ID test} & \textbf{OOD validation} & \textbf{OOD test} \\
        \hline
        \multirow{3}{*}{GOOD-ZINC} & covariate & 149674 & 24945 & 24945 & 24945 & 24946 \\
                                   & concept   & 101867 & 21828 & 21828 & 43539 & 60393 \\
                                   & no shift  & 149673 & 49891 & 49891 & - & - \\
        \hline
    \end{tabular}
    \label{fig:zinc details}
    \end{table}

\subsection{Experimental Settings}
\label{appendix:settings}
We use the GOOD-ZINC dataset from the GOOD benchmark and the S-OOD tasks from ReactionOOD, excluding other OOD tasks from ReactionOOD as they are still under maintenance. Our baseline results on ReactionOOD have been acknowledged by the original authors.
\begin{figure}[h]
    \centering
    \includegraphics[trim=0 0 0 0,clip,width=0.58\textwidth]{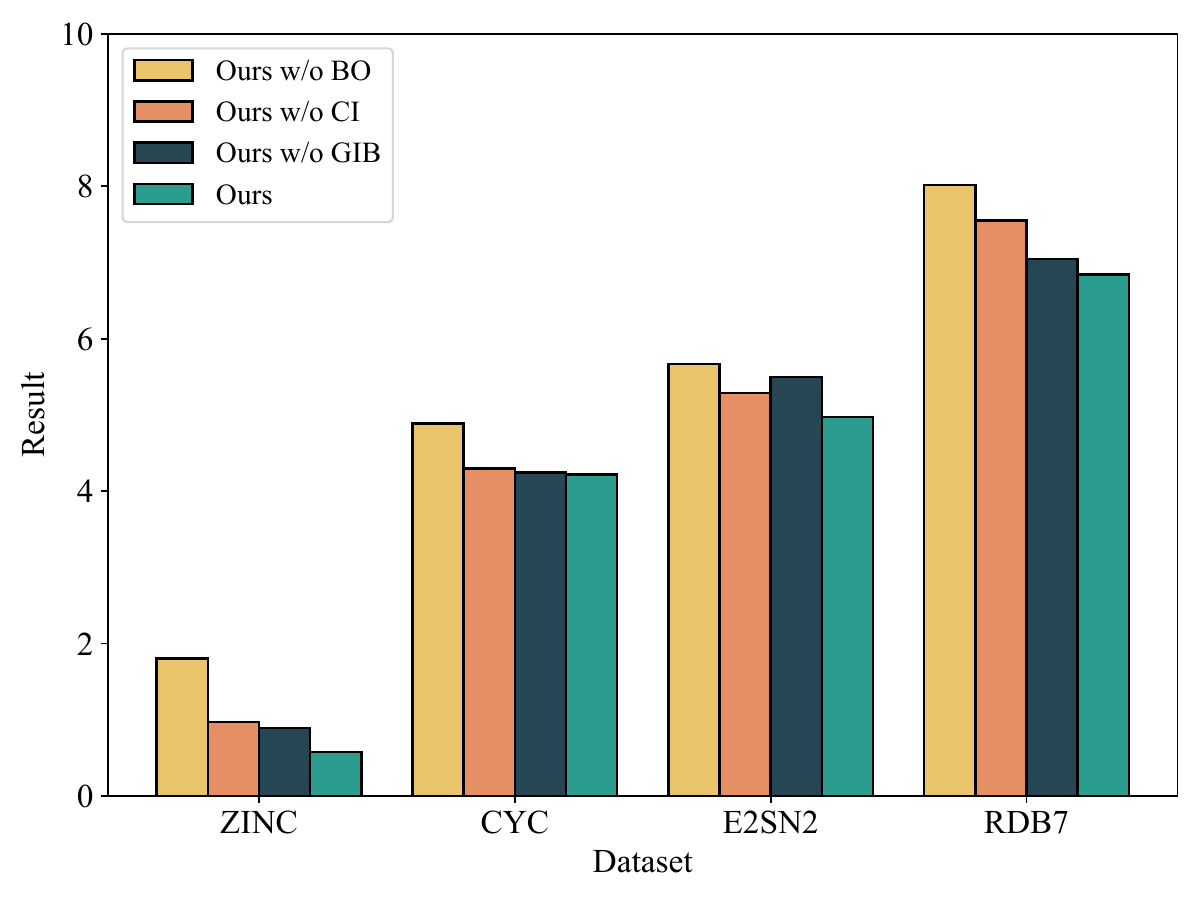}
    \caption{The comparison of different components.}
    \label{fig:loss effectiveness analysis}
\end{figure}
We use a three-layer GIN as the backbone model, with 300 hidden dimensions, which is consistently applied in both OURS and baseline models. The model is trained for 300 epochs, with the learning rate adjusted using the cosine annealing strategy. The initial learning rate is set to 0.001, with a minimum value of 1e-8 . For the OURS model, all tunable hyperparameters in the loss function $L$ are set to 0.5.

\subsection{Ablation Studies}
\label{appendix:ablation}

\paragraph{Effectiveness Analysis}
To evaluate the effectiveness of the proposed loss functions $L_{GIB}$ and $L_{CI}$ in improving the model’s OOD generalization ability, we conducted a series of ablation studies across four ood datasets: ZINC, Cycloaddition, E2SN2, and RDB7. Ours w/o BO serves as the baseline model, where both loss functions are removed, and only the causal subgraph readout layer’s $l_1$ loss is used for optimization. Ours w/o GIB ablates $L_{GIB}$, eliminating the constraint on confounding subgraphs to assess the impact of removing confounder control on generalization. Conversely, Ours w/o CI removes $L_{CI}$ while keeping $L_{GIB}$, allowing us to examine the contribution of the causal intervention loss to OOD generalization. Ours represents the complete model, incorporating both loss functions for optimization. Notably, ZINC is evaluated using MAE, while the other datasets adopt RMSE as the evaluation metric. Given that the ZINC results are small (approximately 0.0x), we scale them by a factor of 10 in the \Cref{fig:loss effectiveness analysis} for better visualization and comparison.

The results reveal several key insights. 
The full model (green) consistently achieves the lowest RMSE across all datasets, demonstrating the effectiveness of jointly applying both the enhanced GIB loss and the CI loss. Removing both components (yellow) leads to the worst performance, confirming that both components are essential. Between the two losses, removing CI (orange) generally causes a larger degradation than removing GIB (blue), suggesting that CI plays a more dominant role. On E2SN2, however, GIB contributes more significantly. These results indicate that GIB and CI provide complementary benefits, and that using both yields the best OOD generalization.

% First, compared to the baseline model (Ours w/o BO), both Ours w/o CI (which retains $L_{GIB}$) and Ours w/o GIB (which retains $L_{CI}$) show notable improvements across all datasets, indicating that both loss functions contribute positively to enhancing OOD generalization. Furthermore, a comparison between Ours w/o GIB and Ours w/o CI demonstrates that retaining $L_{CI}$ generally yields better performance than retaining $L_{GIB}$ alone. This suggests that $L_{CI}$, through causal intervention, enhances the model’s robustness and mitigates the degradation caused by confounding factors to some extent. However, the best performance is consistently achieved by the full model (Ours), indicating that the two losses are complementary and jointly improve OOD generalization.

\paragraph{Parameter Sensitivity Analysis}

In this experiment, we analyzed the sensitivity of loss function hyperparameters under different settings in the Cycloaddition dataset, focusing on two key components of our proposed loss function: the hyperparameter $\lambda$ for the causal intervention term and $\alpha$, $\beta$ for the confounding constraint term.
\begin{figure}[h]
    \centering
    \includegraphics[trim=0 0 0 0,clip,width=0.88\textwidth]{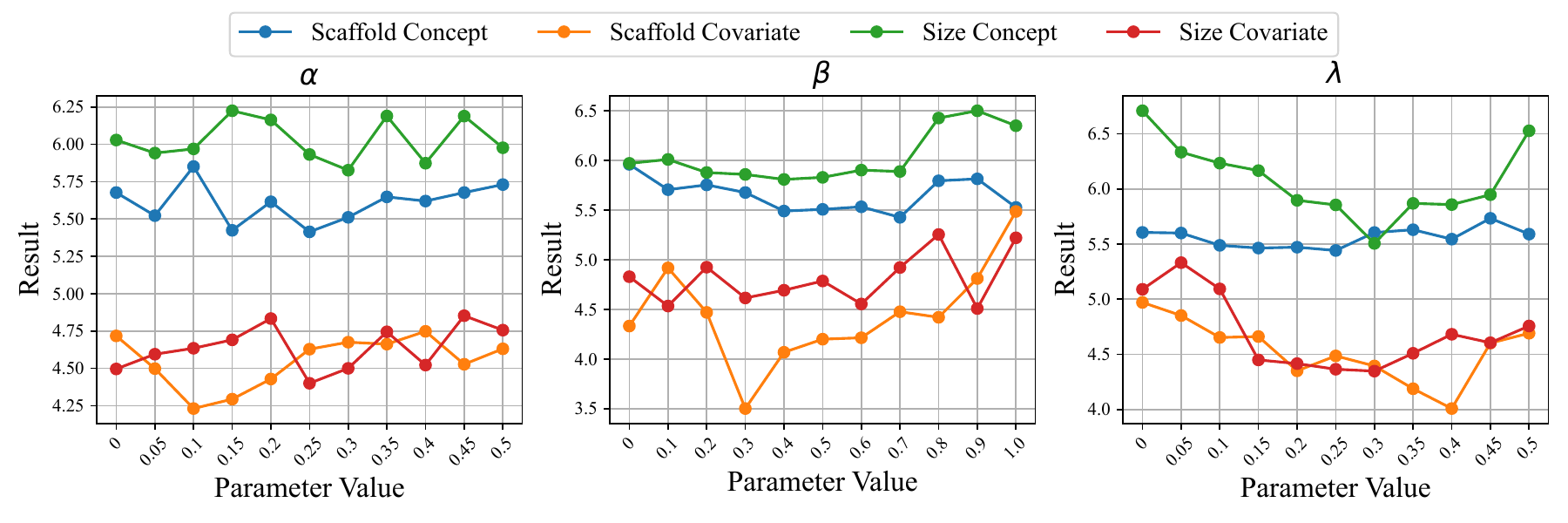}
    \caption{Parameter sensitivity.}
    \label{fig:Parameter sensitivity}
\end{figure}
The results in \Cref{fig:Parameter sensitivity} indicate that, there is no clear trend toward getting better or worse for $\alpha$. For $\beta$, which balances the GIB loss, there is a gradual increase in RMSE when it is too large, especially in scaffold-covariate settings, suggesting an optimal range around 0.3–0.6.  For $\lambda$, which controls the causal intervention loss, has the strongest impact. A suitable parameter interval (0.2–0.4) consistently leads to lower RMSE, while overly large or small $\lambda$ causes performance degradation, especially in the size-concept setting. This demonstrates the importance of carefully tuning $\lambda$ to achieve effective OOD generalization.

\section{Framework Details}
\label{appendix:framework}
Given a GNN-based encoder $f(\cdot)$ and a graph $G_i = \left(A_i, X_i \right)$, the graph representation is computed as:
\begin{equation}
H_{g,i} = f(A_i, X_i),
\end{equation}

Then, to estimate attention scores, inspired by ~\citep{sui2022causal}, we utilize separate MLPs for nodes and edges. The node attention scores, which can be seen as the node-level soft mask can be computed as:
\begin{equation}
\mathrm{M}_{\mathrm{node}}, \bar{\mathrm{M}}_\mathrm{node} = \sigma(\mathrm{MLP}_\mathrm{node}(H_{g,i})),
\end{equation}
where $\sigma$ denotes the softmax operation applied across attention dimensions. Similarly, edge-level soft masks are determined by concatenating node embeddings from connected edges, followed by an edge-specific MLP:
\begin{equation}
\mathrm{M}_{\mathrm{edge}}, \bar{\mathrm{M}}_\mathrm{edge} = \sigma(\mathrm{MLP}_\mathrm{edge}([H_{g,i}[\mathrm{row}],H_{g,i}[\mathrm{col}]])),
\end{equation}

These soft masks serve as weighting mechanisms, allowing the model to focus on the most relevant nodes and edges while maintaining differentiability.

Next, we decompose the initial graph to causal and confounding attened-subgraph:
\begin{equation}
\text{C}_i= \left\{A_i \odot M_{edge}, X_i \odot M_{node} \right\},
\end{equation}
\begin{equation}
    \text{S}_i= \left\{A_i \odot \bar{M}_{edge}, X_i \odot \bar{M}_{node} \right\}.
\end{equation}

To encode these subgraphs, $C_i$ and $S_i$ are processed through a pair of GNNs with shared parameters, extracting causal and confounding representations $H_c$ and $H_s$, respextively. Finally, the representations of the two subgraphs are respectively used to obtain the predictions of the regression task through the corresponding readout layers.

\section{Variational Bounds for the GIB Objective}
The mutual information $I(C; G)$ quantifies the dependency between $C$ and $G$ and is defined as:
\begin{equation}
I(C; G) = \mathbb{E}_{p(C, G)} \left[ \log \frac{p(C \mid G)}{p(C)} \right]. \label{eq:mutual_info}
\end{equation}

However, computing the marginal distribution $p(C) = \sum_G p(C \mid G) p(G)$ is intractable, to overcome this challenge, we approximate $p(C)$ with a variational distribution $q(C)$. Substituting $q(C)$ into Eq.~\eqref{eq:mutual_info}, we reformulate $I(C; G)$ as:
\begin{equation}
I(C; G) = \mathbb{E}_{p(C, G)} \left[ \log \frac{p(C \mid G)}{q(C)} \right] - \mathrm{KL}\big(p(C) \| q(C)\big). \label{eq:variational_mi}
\end{equation}

The KL divergence term $\mathrm{KL}\big(p(C) \| q(C)\big)$ is non-negative, providing an upper bound for $I(C; G)$:
\begin{equation}
I(C; G) \leq \mathbb{E}_{p(G)} \big[ \mathrm{KL}\big(p(C \mid G) \| q(C)\big) \big]. \label{eq:mi_upper_bound_app}
\end{equation}

%%%%%%%%%%%%%%%%%%%%%%%%%%%%%%%%%%%%%%%%%%%%%%%%%%%%%%%%%%%%%%%%%%%%%%%%%%%%%%%
%%%%%%%%%%%%%%%%%%%%%%%%%%%%%%%%%%%%%%%%%%%%%%%%%%%%%%%%%%%%%%%%%%%%%%%%%%%%%%%

\end{document}